%% file: cas-dc-template.tex
\documentclass[a4paper,fleqn,longmktitle]{cas-dc}
\usepackage[numbers]{natbib}
\usepackage[T1,T2A]{fontenc}
\usepackage[utf8]{inputenc}
\usepackage[english]{babel}
\usepackage{diagbox}

\def\tsc#1{\csdef{#1}{\textsc{\lowercase{#1}}\xspace}}
\tsc{WGM}
\tsc{QE}
\tsc{EP}
\tsc{PMS}
\tsc{BEC}
\tsc{DE}
\newcommand{\Fp}{F_1^\text{partial}}
\newcommand{\Fe}{F_1^\text{exact}}

\begin{document}
\let\WriteBookmarks\relax
\def\floatpagepagefraction{1}
\def\textpagefraction{.001}
\shorttitle{Russian language corpus with a developed deep learning neuronet complex to analyze it}
\shortauthors{SG Sboeva et~al.}

\title [mode = title]{An analysis of full-size Russian complexly NER labelled corpus of Internet user reviews on the drugs based on deep learning and  language neural nets }     




\author[1,2]{AG Sboev}[type=editor, orcid=0000-0002-6921-4133]
\cormark[1]
\ead{sag111@mail.ru}
\author[3]{SG Sboeva}
\ead{sboevasanna@mail.ru}

\credit{Conceptualization of this study, Methodology, Dataset}

\address[1]{NRC "Kurchatov institute", Moscow, Russia}
\address[2]{MEPhI National Research Nuclear University, Kashirskoye sh., 31, Moscow, 115409, Russia}
\address[3]{I.M. Sechenov First Moscow State Medical University (Sechenov University), Moscow, Russia}


\author[1]{IA Moloshnikov}
\ead{ivan-rus@yandex.ru}

\author[1]{AV Gryaznov}
\ead{artem.official@mail.ru}

\author[1]{RB Rybka}
\ead{rybkarb@gmail.com}

\author[1]{AV Naumov}
\ead{sanya.naumov@gmail.com}

\author[1]{AA Selivanov}
\ead{aaselivanov.10.03@gmail.com}

\author[1]{GV Rylkov}
\ead{gvrylkov@mail.ru}


\author[1]{VA Ilyin}
\ead{ilyin0048@gmail.com }




\cortext[cor1]{Corresponding author}
\sloppy 
\begin{abstract}
We present the full-size Russian complexly NER-labeled corpus of Internet user reviews, along with an evaluation of accuracy levels reached on this corpus by a set of advanced deep learning neural networks to extract the  pharmacologically meaningful entities from Russian texts. The corpus annotation includes mentions of the following entities: Medication (33005 mentions), Adverse Drug Reaction (1778), Disease (17403), and Note (4490). Two of them – Medication and Disease – comprise a set of attributes. A part of the corpus has the coreference annotation with 1560 coreference chains in 300 documents. 
Special multi-label model based on a language model and the set of features is developed, appropriate for presented corpus labeling. The influence of the choice of different modifications of the models: word vector representations, types of language models pre-trained for Russian, text normalization styles, and other preliminary processing are analyzed. The sufficient size of our corpus allows to study the effects of particularities of corpus labeling and balancing entities in the corpus. As a result, the state of the art for the pharmacological entity extraction problem for Russian is established on a full-size labeled corpus.
In case of the adverse drug reaction (ADR) recognition, it is 61.1 by the F1-exact metric that, as our analysis shows,  is on par with the accuracy level  for other language corpora  with similar characteristics and the ADR representativnes. The evaluated baseline precision of coreference relation extraction on the corpus is 71, that is higher the results reached on other Russian corpora.

\end{abstract}




\maketitle
\section{Introduction}
Nowadays, a great amount of texts collected in the open Internet sources contains a vast variety of socially significant information. In particular, such information relates to healthcare in general, consumption sphere and evaluation of medicines by the population. Due to time limitations, clinical researches may not reveal the potential adverse effects of a medicine before entering the pharmaceutical market. This is a very serious problem in healthcare. Therefore, after a pharmaceutical product comes to the market, pharmacovigilance (PV) is of great importance. Patient opinions on the Internet, in particular in social networks, discussion groups, and forums, may contain a considerable amount of information that would supplement clinical investigations in evaluating the efficacy of a medicine. Internet posts often describe adverse reactions in real time ahead of official reporting, or reveal unique characteristics of undesirable reactions that differ from the data of health professionals. Moreover, patients openly discuss a variety of uses of various drugs to treat different diseases, including ``off-label'' applications. This information would be very useful for a PV database where risks and advantages of drugs would be registered for the purpose of safety monitoring, as well as the possibility to form hypotheses of using existing drugs for treating other diseases. This leads to an increasing need for the analysis of Internet information 
to assess the quality of medical care and drug provision. In this regard, one of the main tasks is the development of machine learning methods for extracting useful information from social media. 
However, expert assessment of such amount of  text information is too laborious, therefore special methods 
 have to be developed with taking into account the presence in  these texts the informal vocabulary and  of reasoning.  
The quality of these methods directly depends on tagged corpora to train them. 
In this paper, 
we present the full-size Russian complexly NER-labeled corpus of Internet user reviews, named Russian Drug Reviews corpus of SagTeam project (RDRS)\footnote{Corpora description is presented on \url{https://sagteam.ru/en/med-corpus/} }-  comprising the part with tagging on coreference relations. 
Also, we present model appropriate to the corpus multi-tag labelling  developed on base of the combination of XLM-RoBERTa-large model with the set of added features.

 In Section~\ref{sec:related_works}, we analyse the selected set of corpora comprising ADR (Adverse drug reaction) labels, but different by  fillings, labeling tags, text sizes and styles with a goal to analyse their influence on the ADR extraction precision.
The materials used to collect the corpus are outlined~in~Section~\ref{sec:corpus_collecting}, the technique of its annotation is described in Section~\ref{subsec:corpus_annotation}. The developed machine learning complex is presented in Section~\ref{sec:Methods}. The conducted numerical experiments are presented in Section~\ref{subsec:Experiments} and discussed in~Sections~\ref{sec:results}~and~\ref{sec:discussion} .

\section{Related works}
\label{sec:related_works}
 In world science, research concerning the above-mentioned problems is conducted intensively, resulting in a great diversity of annotated corpora. From the linguistic point of view, these corpora can be distinguished  into two groups: firstly, the ones of texts written by medics (clinical reports with annotations), and secondly, those of texts written by non-specialists, namely, by the Internet customers who used the drugs. The variability of the natural language constructions in the speech of Internet users complicates the analysis of corpora based on Internet texts, вut there are the other distinctive features of any corpus : the number of entities, the number of annotated phrases definite types, also the number of its mutual uses in phrases, and approaches to entity normalization.  The diversity of these features influences the accuracy of entity recognition on the base of different corpora. Also the  tipes of entity labelling  and   used metrics of evaluating results may be various.   Not for each corpus  a necessary information is available.
 Below we briefly describe 6 corpora: CADEC, n2c2-2018, Twitter annotated corpus, PsyTAR, TwiMed corpus, RuDReC. 

\subsection{Corpora description}

\paragraph{CADEC (corpus of adverse drug event annotations)
~\cite{karimi2015cadec}} is a corpus of medical posts taken from the AskaPatient~\footnote{Ask a Patient: Medicine Ratings and Health Care Opinions - \url{http://www.askapatient.com/}}
forum and annotated by medical students and computer scientists. It collects ratings and reviews of medications from their consumers and contains consumer posts on 13 different drugs. There are 1253 posts with 7398 sentences. The following entities were annotated: Drug, ADR, Symptom, Disease, Findings. The annotation procedure involved 4 medical students and 2 computer scientists. In order to coordinate the markup, all annotators jointly marked up several texts, and after that the texts were distributed among them. All the annotated texts were checked by three corpus authors for obvious mistakes, e.g. missing letters, misprints, etc.

\paragraph{TwiMed corpus (Twitter and PubMed comparative corpus of drugs, diseases, symptoms, and their relations)}~\cite{alvaro2017twimed} contains 1000 tweets and 1000 sentences from Pubmed~\footnote{National Center for Biotechnology Information webcite - \url{http://www.ncbi.nlm.nih.gov/pubmed/}}
for 30 drugs. It was annotated for 3\,144 entities, 2\,749 relations, and 5\,003 attributes. The resulting corpus was composed of agreed annotations approved by two pharmaceutical experts. The entities marked were Drug, Symptom, and Disease.

\paragraph{Twitter annotated corpus~\cite{sarker2016social}} consists of randomly selected tweets containing drug name mentions: generic and brand names of the drugs. The annotator group comprised pharmaceutical and computer experts. Two types of annotations are currently available: Binary and Span. The binary annotated part~\cite{sarker2015portable} consists of 10\,822 tweets annotated by the presence or absence of ADRs. Out of these, 1\,239 (11.4\%) tweets contain ADR mentions and 9583 (88.6\%) do not. 
The span annotated part~\cite{sarker2016social} consists of 2\,131 tweets (which include 1\,239 tweets containing ADR mention from the binary annotated part). The semantic types marked are: ADR, beneficial effect, indication, other (medical signs or symptoms). 

\paragraph{PsyTAR dataset~\cite{zolnoori2019psytar}} contains 891 reviews on four drugs, collected randomly from an online healthcare forum~\footnote{Ask a Patient: Medicine Ratings and Health Care Opinions - \url{http://www.askapatient.com/}}
. They were split into 6\,009 sentences. To prepare the data for annotation, regular expression rules were formulated to remove any personal information such as emails, phone numbers, and URLs from the reviews. The annotator group included pharmaceutical students and experts. They marked the following set of entities: ADR, Withdrawal Symptoms (WD), Sign, Symptom, Illness (SSI), Drug Indications (DI) and other. Sadly, the original corpus doesn't contain mentions boundaries in source texts. It complicates the NER task. In a paper \cite{basaldella2019bioreddit} presented version of the PsyTAR corpus in CoNLL format, where every word has corresponding tag of named entity. We use this version for comparison purposes.

\paragraph{n2c2-2018~\cite{n2c2-2018}}
 is a dataset from the National NLP Clinical Challenge of the Department of Biomedical Informatics (DBMI) at Harvard Medical School. 
The dataset contains clinical narratives, and builds on past medication extraction tasks, but examines a broader set of patients, diseases, and relations as compared with earlier challenges. 
It was annotated by 4 paramedic students and 3 nurses. Label set includes medications and associated attributes, such as dosage (Dosage), strength of the medication (Strength), administration mode (Mode), administration frequency (Frequency), administration duration (Duration), reason for administration (Reason), and drug-related adverse reactions (ADEs).
The number of texts was 505 (274 in training, 29 in development and 202 in test).

\paragraph{RuDReC~\cite{10.1093/bioinformatics/btaa675}}
Labeled part of RuDReC contains 500 reviews on drugs from a medical forum OTZOVIK. Two step annotation procedure was performed: on first step authors used 400 texts labeled according formats of site Sagteam [https://sagteam.ru/en/med-corpus/annotation/] by 4 experts of Sechenov First Moscow State Medical University - now participants of our projects; on second step they simplified labeling by deleting/uniting tags and annotated in addition 100 reviews. Totally in RuDReC and in proposed corpus RDRS 467 texts are coincident. An influence of differences in labelling of them on the ADR extraction accuracy presented in Section~\ref{sec:discussion}.

\subsection{Target vocabularies in the corpora normalization}
The normalization task of internet user texts is more difficult because of informal text style and more natural vocabulary. Still, as in the case of clinical texts, thesauruses are used. In particular, annotated entities in CADEC were mapped to controlled vocabularies: SNOMED CT, The Australian Medicines Terminology (AMT)~\cite{techrepAMT}, and MedDRA. Any span of text annotated with any tag was mapped to the corresponding vocabularies. If a concept did not exist in the vocabularies, it was assigned the ``concept\_less'' tag. In the TwiMed corpus, for Drug entities the SIDER database~\cite{kuhn2015sider} was used, which contains information on marketed medicines extracted from public documents, while for Symptom and Disease entities the MedDRA ontology was used. In addition, the  terminology of SNOMED CT concepts was used for entities, which belong to the Disorder semantic group. In the Twitter dataset~\cite{sarker2016social}, when annotating ADR mentions, they were set in accordance to their UMLS concept ID. Finally, in PsyTAR corpus, ADRs, WDs, SSIs and DIs entities were matched to UMLS Metathesaurus concepts and SNOMED CT concepts. No normalizations was applied to n2c2-2018 corpus. 

\subsection{Number of entities and their breakdown in the corpora}
In Table \ref{tab:adr_saturation}, we review the complexity characteristics of the selected corpora and evaluate the dependence of accuracy of extracting the ADR on them.
The overlap entities  are  only in few  of considered corpora but   their parts are relatively small, excluding CADEC, where there are the parts of overlap ADR entities, both continuous (5\%), and discontinuous (9\%). 
In this sense, CADEC, appears, is the most complicated corpus from selected, but having the largest numbers of ADR mentions and the largest value of the relation of ADR mention number to symptom  mention number. If the first factor complicates  the ADR identification, both others simplify. We could not find in literature the information  about  the  precision of the ADR  identification for all  corpora in view according metrics exact F1. However, on the base of data of Table \ref{tab:adr_saturation} we suggest the parameter of relation of the ADR mention number  to total number of corpus words is convenient to compare  the corpora, and we use it further named as "saturation".


\subsection{Coreference task}
There is a problem, that some reviews present user opinion concerning the mentions of a particular tag in relation to more than one entity of real world: a drug, or disease, or the other entities. For example, some reviews may contain reports about use of multiple medications that may have different effects, so coreference annotation may be useful for detection of different mentions referred to the same drug. For English language there are few corpora for coreference resolution like CoNLL-2012~\cite{pradhan2012conll} or GAP~\cite{webster2018gap}, and even corpus of pharmacovigilance records with adversarial drug reactions annotations that includes coreference annotation (PHAEDRA)~\cite{thompson2018annotation}. The coreference problem  in Russian texts is slightly highlighted in a literature. Currently, there are only two corpora with coreference annotations for Russian language: Ru-Cor~\cite{azerkovich2014evaluating6887693} and corpus from shared task AnCor-2019~\cite{Ju2014RUEVAL2019EA}. The latter is a continuation and extension of the first. As for the methods the state-of-the-art approach is based on neural network trained end-to-end to solve two task at the same time: mention extraction and relations extraction. This approach was firstly introduced in \cite{lee2017end} and have been used in several papers\cite{lee2018higher, joshi2019bert, xu2020revealing, joshi2020spanbert, toshniwal2020learning} with some modifications to get higher scores on the coreference corpus CoNLL-2012~\cite{pradhan2012conll}.


\begin{table*}
\caption{A sample post for ``Глицин'' (Glycine) from otzovik.com. Original text is quoted, and followed by English translation in parentheses.}
\label{tab:sample_post_MEDcorp}
\input{tables/table5.tex}
\end{table*}

\begin{table*}
\centering
\caption{Numerical estimation of the corpora complexity on ADR level saturation. \newline Explanation of abbreviations of corpora names: TA – Twitter Annotated Corpus, TT – TwiMED Twitter, TP – TwiMED PubMed, N2C2 – n2c2-2018. Following the artcile \cite{gupta2018co}, we meant by the ADRs symptoms related to the drugs in TT and TP corps.  Explanation of abbreviations of metrics: f1-e – f1-exact, f1-am – f1-approximate match,  f1-r – f1-relaxed, f1-cs f1 - Classification of sentences with ADR, NA - data not available for download and analysis }
\input{tables/adr_saturation.tex}
\label{tab:adr_saturation}
\end{table*}

\begin{table*}
\centering
\caption{Proportions of difficult cases in annotations. Discontinuous mentions are labeled phrases separated by words not related to it. A mention is overlapping if some of its words also labeled as another mention.}
\input{tables/types_saturation.tex}
\label{tab:types_saturation}
\end{table*}

\section{Corpus collecting} \label{sec:corpus_collecting}
\subsection{Corpus material} \label{subsec:corpus_material}
 

In this section, we report  the design of our corpus. Its basis were 2\,800 reviews from a medical section of the forum called OTZOVIK\footnote{OTZOVIK - Internet forum from which user reviews were taken - 
\url{http://otzovik.com}}
, which is dedicated to consumer reviews on medications. On that website there is a partition where users submit posts by filling special survey forms. The site offers two forms: simplified and extended, the latter being optional. In this form a user selects a drug name and fills out the information about the drug, such as: adverse effects experienced, comments, positive and negative sides, satisfaction rate, and whether they would recommend the medicine to friends. In addition, the extended form contains prices, frequency, scores on a 5-point scale for such parameters as quality, packing, safety, availability. A sample post for ``Глицин'' (Glycine) is shown in Table~\ref{tab:sample_post_MEDcorp}.

We used information only from the simplified form, since the users had rarely filled extended forms in their reviews. We considered only the fields Heading, General impression and Comment. Furthermore, some of the reviews are written in common language and do not follow formal grammar and punctuation rules. The consumers described not only their personal experience, but sometimes opinions of their family members, friends or others. 

\subsection{Corpus Annotation} \label{subsec:corpus_annotation}
This section describes the corpus annotation methodology, including the markup composition, the annotation procedure with guidelines for complex cases, and software infrastructure for the annotation.

\subsubsection{Annotation process}\label{subsec:annotation_process}
The group of 4 annotators annotated  review texts using a guide developed jointly by machine learning experts and pharmacists. Two annotators were certified pharmacists, and the two others were students with pharmaceutical education. Reliability was achieved through joint work of annotators on the same set of documents, subsequently controlled by means of journaling. After the initial annotation round, the annotations were corrected three times with cross-checking by different annotators, after which the final decision was made by an expert pharmacist.
The corpus annotation comprised the following steps:
\begin{enumerate}
    \item First, a guide was compiled for the annotators. It included entities description and examples.
    \item Upon testing on a set of 300 reviews, the guide was corrected, addressing complex cases. During that, iterative annotation was performed, from 1 to 5 iterations for a text, while tracking for each text and each iteration the annotator questions, controller comments, and correction status.
    \item The resulting guide was used for annotating the remaining reviews. Two annotators marked up each review, and then a pharmacist checked the result. When complex cases were found, they were analyzed separately by the whole group of experts.
    \item The obtained markup was automatically checked for any inaccuracies, such as incomplete fragments of words selected as mentions, terms marked differently in different reviews, etc. Texts with such inaccuracies were rechecked.
\end{enumerate}

 To estimate an agreement between annotators   
14
we used the metric described by Karimi et al.  \cite{karimi2015cadec}.  According to this metric we calculated the agreement score for every document as the ratio between number of matched mentions and maximum number of mentions, annotated by one of the annotators in current document. Matched mentions are calculated depending on two flags $\alpha$ and $\beta$. The first one is the span strictness, it can be \textit{strict} or \textit{intersection}. If we do a strict spans comparison then only mentions with equal borders will be counted as matching, otherwise, we count mentions as matching if they at least are intersected each other. But every mention annotated by each annotator can be matched with the only mention annotated by the other annotator. $\beta$   is the tag strictness argument, which can be \textit{strict} or \textit{ignored}. It defines if we count matched mentions only when both annotators labeled them identically, or we count matched mentions only by borders, despite of labels. After calculation of agreement scores for all documents, we calculate the average score of the total agreement between two annotators. The average pairwise agreement among annotators is presented in table  \ref{tab:agreement}.
$$\mathrm{agreement}(i,j) = 100\frac{\mathrm{match}(A_i,A_j,\alpha,\beta)}{\mathrm{max}(|A_i|,|A_j|)}$$
Here $A_i$ and $A_j$ are lists of mentions annotated by annotators $i$ and $j$. $|A_i|$ and $|A_j|$ are numbers of elements in these lists.
\begin{table}
\caption{Average pair-wise agreement  between annotators} 
\label{tab:agreement} 
\input{tables/agreement_t.tex} 
\end{table}

The annotation was carried out with the help of the WebAnno-based toolkit, which is an open source project under the Apache License v2.0.
It has a web interface and offers a set of annotation layers for different levels of analysis.  Annotators proceeded according to the guidelines below.

\subsubsection{Guidelines applied in the course of annotation}
\label{subsec:Guidelines}
The annotation goal was to get a corpus of reviews in which named entities reflecting pharmacotherapeutic treatment are labelled, and annotate medication characteristic semantically. With this in mind, the objects of annotation were attributes of drugs, diseases (including their symptoms), and undesirable reactions to those drugs. The annotators were to label mentions of these three entities with their attributes defined below.

\paragraph{Medication.} This entity includes everything related to the mentions of drugs and drugs manufacturers. Selecting a mention of such entity, an annotator had to specify an attribute out of those specified in Table~\ref{tab:medication}, thereby annotating it, for instance, as a mention of the attribute ``DrugName'' of the entity ``Medication'' .
In addition, the attributes ``DrugBrand'' and ``MedFrom'' were annotated with the help of lookup in an external source~\cite{SRD}.

\begin{table*}
\caption{Attributes belonging to the Medication entity}
\label{tab:medication}
\input{tables/medication.tex}
\end{table*}

\paragraph{Disease.} 
This entity is associated with diseases or symptoms. It indicates the reason for taking a medicine, the name of the disease, and improvement or worsening of the patient state after taking the drug. 
Attributes of this entity are specified in Table~\ref{tab:Disease}.

\begin{table*}
\caption{Attributes belonging to the Disease entity} 
\label{tab:Disease} 
\input{tables/disease.tex}
\end{table*}

\paragraph{ADR.} This entity is associated with adverse drug reactions in the text. For example, one post said:
<<После недели приема Кортексина у ребенка начались судороги>> (After a week of taking Cortexin, the child began to cramp). In this sentence, the word ``судороги'' (``cramp'') is labeled as an ADR entity.

\paragraph{Note.} We use this entity when the author makes recommendations, tips, and so on, but does not explicitly state whether the drug helps or not. These include phrases like ``I do not advise''. For instance, the phrase <<Нет поддержки для иммунной системы>> (No support for the immune system) is annotated as a Note.

\begin{figure}
    \centering
    \includegraphics[width=8.2cm]{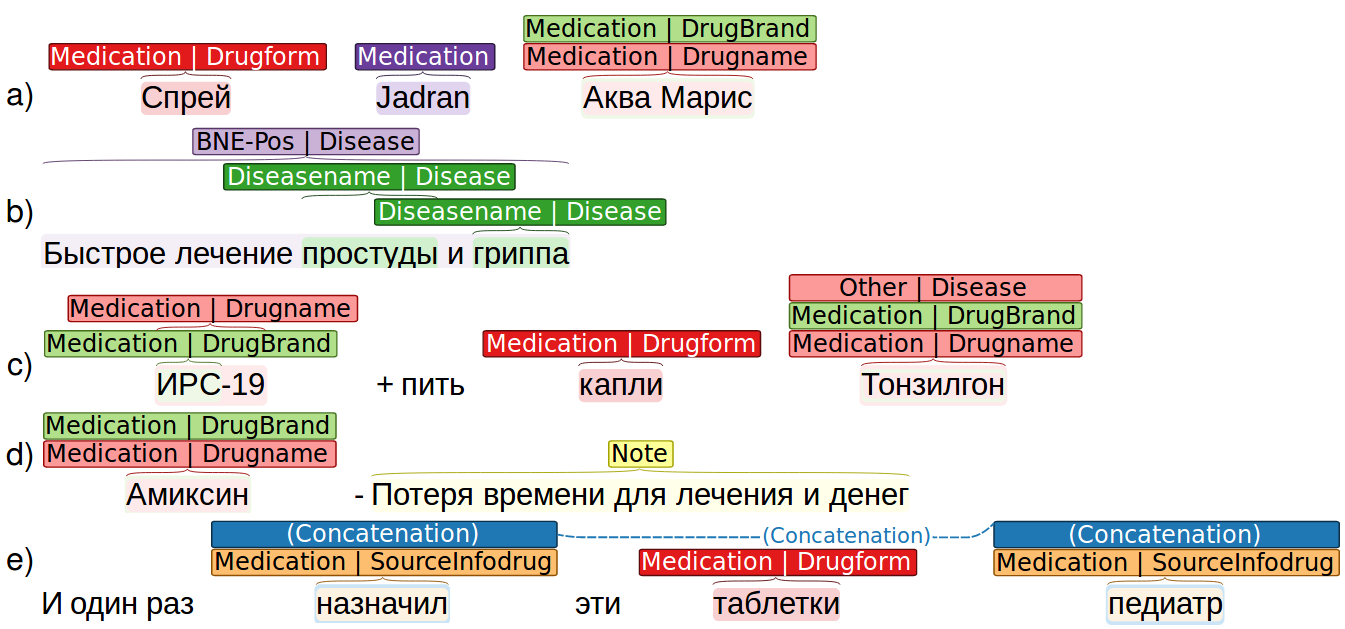}
    \caption{Examples of markup. a) ``Spray Jadran Aqua Maris'', b) ``Rapid treatment of cold and flu'', c) ``IRS-19 + drink drops of Tonsilgon'' d) ``Amixin -- waste of time and money for treatment'', e) ``And once were these pills prescribed by my pediatrician''}
    \label{fig:webannoExample1}
\end{figure}

The typical situations that had to be handled during the annotation are the following:
\begin{enumerate}
    \item A simple markup, when a mention consists of 1 or more words and it related to a single attribute of entity. The annotators then just have to select a minimal but meaningful text fragment, excluding conjunctions, introductory words, and punctuation marks.
    \item Discontinuous annotation -- when mentions separated by words that are not part of it. It is then necessary to annotate mention parts and connect them. In such cases we use the ``concatenation'' relation. In the example (e) on Fig.~\ref{fig:webannoExample1} the words ``prescribed'' and ``pediatrician'' are annotated as a concatenated parts of mention of the attribute ``sourceInfoDrug''.
    \item Intersecting annotations. Words in a text can belong to mentions of different entities or attributes simultaneously. For example, in the sentence ``Rapid treatment of cold and flu'' (see Fig.~\ref{fig:webannoExample1}, example (b)), words ``cold'' and ``flu'' are mentions of attribute ``diseasename'', but at the same time the whole phrase is a mention of attribute ``BNE-Pos''. If a word or a phrase belongs to a mentions of different attributes or entities at the same time (for example, ``drugname'' and ``drugbrand''), it should be annotated with all of them: see, for instance, entity ``Aqua Maris'' in sentence ``Spray Jadran Aqua Maris'' (Fig.~\ref{fig:webannoExample1}, example (a)).
    \item Another complex situation is when an analogue (or, in some cases, several analogues) of the drugs are mentioned in a text, for example, when a customer wrote about a drug and then described an alternative that helped them. In this case, the ``Other'' attribute is used (example (c)).
\end{enumerate}


Moreover, there often were author subjective arguments instead of explicit reports on the outcomes. We labeled that as a mention of entity ``Note''. For example, ``strange meds'', ``not impressed'', ``it is not clear whether it worked or not'', ``ambiguous effect'' (example (d) in Fig.~\ref{fig:webannoExample1}). 

\subsection{Classification based on categories of the ATC, ICD-10 classifiers and MedDRA terminology}
    \label{subsec:Normalization}
After annotation, in order to resolve possible ambiguity in terms we performed normalization and classification by matching the labeled mentions to the information from external official classifiers and registers. The external sources for Russian are described below.

\begin{itemize}
    \item 
    the 10-th revision of the International Statistical Classification of Diseases and Related Health Problems (ICD-10)~\cite{ICD-10} is an international classification system for diseases which includes 22 classes of diagnoses, each consisting of up to 100 categories. The ICD-10 makes it possible to reduce verbal diagnoses of diseases and health problems to unified codes.
    \item The Anatomical Therapeutic Chemical (ATC)~\cite{miller1995new} is an international medication classification system containing 14  anatomical main groups and 4 levels of subgroups. The ICD-10 and the ATC have a hierarchical structure, where ``leaves'' (terminal elements) are specified diseases or medications, and ``nodes'' are groups or categories. Every node has a code, which includes the code of its parent node.
    \item State Register of Medicinal Products (SRD)(``Государственный реестр лекарственных средств (ГРЛС)''~\cite{SRD} in Russian) is a register of detailed information about the medications certified in the Russian Federation. It includes possible manufacturers, dosages, dosage forms, ATC codes, indications, and so on.
    \item MedDRA\textregistered{}  the Medical Dictionary for Regulatory Activities terminology is the international medical terminology developed under the auspices of the International Council for Harmonisation of Technical Requirements for Pharmaceuticals for Human Use (ICH)
\end{itemize}

Among the international systems of standardization of concepts, the most complete and large metathezaurus is UMLS, which combines most of the databases of medical concepts and observations, including MESH (and MESHRUS), ATC, ICD-10, SNOMED CT, LOINC and others. Every unique concept in the UMLS has an identification code CUI, using which one can get information about the concept from all the databases. However, within UMLS it is only the MESHRUS database that contains Russian language and can be used to associate words from our texts with CUI codes. 

Classification was carried out by the annotators manually. For this purpose, we applied the procedure consisting of the following steps:
automatic grouping of mentions, manual verification of mention groups (standardization), matching the mention groups to the groups from the ATC and the ICD-10 or terms from MedDRA.

Automatic mentions grouping is based on calculating the similarity between two mentions by the Ratcliff/Obershelp algorithm~\cite{ratcliff1988pattern}, which is based on searching two strings for matching substrings. In the course of the analysis, every new mention is added to one of the existing groups $G$ if the mean similarity between the mention and all the group items is more than 0.8 (value deduced empirically), otherwise a new group is created. The $G$ set is empty at the start, and the first mention creates a new group with size 1. Each group is named by its most frequent mention. Next, the annotators manually check and refine the resulting set, creating larger groups or renaming them. Mentions of drug names were standardized according to State Register of Medicinal Products. That gave us 550 unique drug names mentioned in corpus.

After that, the group names for attributes ``Diseasename'', ``Drugname'' and ``Drugclass'' are manually matched with ICD-10 and ATC terms to assign term codes from the classifiers. As a result, 247 unique ICD-10 codes were matched against the 765 unique phrases, annotated as attribute ``Diseasename''; 226 unique ATC codes matched the 550 unique drug names; and 70 unique ATC codes corresponded to 414 unique phrases, annotated as ``Drugclass''. Some drug classes that were mentioned in corpus (such as homeopathy) did not have a corresponding ATC code, and were aggregated according to their anatomical and therapeutic classification in the SRD.

Standardized terms for ADR and Indications were manually matched with low level terms (LLT) or prefered terms (PT) from MedDRA. In table~\ref{tab:info_collected_corpus} we show the numbers of Unique PT terms that were matched with our mentions.


\subsection{Statistics of the collected corpus}\label{subsec:Statistics_of_corpus}
We used UDPipe\cite{straka2016udpipe} package to parse the reviews, in order to get sentence segmentation, tokenization and lemmatization. Given this, we calculated that average number of sentences for the reviews is 10, average number of tokens is 152 (with a standard deviation of 44), average number of lemmas is 95 (standard deviation equals to 23). TTR (type/token ratio) was calculated as the ratio of the unique lemmas in a review to the amount of tokens in it. Average TTR for all reviews equals to 0.64.

Detailed information about the annotated corpus is presented in Table~\ref{tab:info_collected_corpus} including:
\begin{enumerate}
    \item The number of mentions for every attribute (``Mentions -- Annotated'' column in the table).
    \item The number of unique classes from classifiers or unique normalized terms described in Section~\ref{subsec:Normalization} matched with our mentions (``Mentions -- Classification \& normalization'').
    \item The number of words belonging to mentions of the attribute (``Mentions -- Number words in the mentions'').
    \item The number of reviews containing any mentions of the corresponding attribute (``Mentions -- Reviews coverage'').
\end{enumerate}

\begin{table*}
\centering
\caption{General information about the collected corpus.} \label{tab:info_collected_corpus}
\input{tables/general_inf_about_corpus_2.tex}
\end{table*}



The corpus contains consumer posts on drugs, mentioned 8 236 times and related to 226 ATC codes.  The most popular 20\% of the ATC codes (by the number of reviews with corresponding Drugname mentions) include 45 different codes which mentions appears in 2\,614 reviews (93\% of all reviews). Among them, 20 ATC codes were reviewed in more then 50 posts (2511 posts in total).

The most popular ATC codes from 2nd level are: L03 ``Immunostimulants'' - 662 reviews (which is 23.6\% of corpus), J05 ``Antivirals for systemic use'' - 508 (18.5\%) reviews, N05 ``Psycholeptics'' - 449 (16.0\%), N02 ``Analgesics'' - 310 (11.1\%), N06 ``Psychoanaleptics'' - 294 (10.5\%). Most popular drugs among immunostimulants by the reviews count are: Anaferon (144 reviews), Viferon (140), Grippferon (71). Most popular antivirals for systemic use are following: Ingavirin (99), Kagocel (71) and Amixin (58).

The proportions of reviews about domestic drugs and foreign to the total number of reviews are 44.9\% and 39.7\% respectively. The remaining documents (15.4\%) contains mentions of multiple drugs both domestic and foreign or mentions of drugs which origin the annotators could not determine. Among the domestic drugs are following: Anaferon (144 reviews), Viferon (140), Ingavirin (99) and Glycine (98). Examples of mentioned foreign drugs: Aflubin (93), Amison (55),  Antigrippin (51) and Immunal (42).



\begin{figure}
    \centering
    \includegraphics[width=8.5cm]{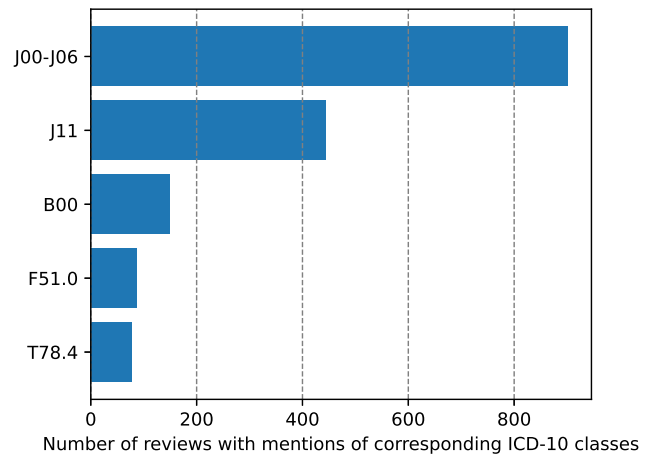}
    \caption{Top 5 low-level disease categories from the ICD-10 by the number of reviews in our corpus. J00-J06 - Acute upper respiratory infections, J11 - Influenza with other respiratory manifestations, virus not identified, B00 - Herpesviral [herpes simplex] infections, F51.0 - Nonorganic insomnia, T78.4 - Allergy, unspecified}
    \label{fig:Top_5_diseases}
\end{figure}

Regarding diseases, the most frequent ICD-10 top level categories are ``X - Diseases of the respiratory system'' (1122 reviews); ``I - Certain infectious and parasitic diseases'' (300 reviews); ``V - Mental and behavioural disorders'' (170 reviews); ``XIX - Injury, poisoning and certain other consequences of external causes'' (82 reviews). The top 5 low level codes from the ICD-10 by the number of reviews are presented in Fig.~\ref{fig:Top_5_diseases}.

Analysing the consumers' motivation to acquire and use drugs (``sourceInfoDrug'' attribute) showed that review authors mainly mention using drugs based on professional recommendations. 989 reviews contains references of doctor prescriptions, 262 - refers to pharmaceutical specialists recommendations and 252 - doctor recommendations. Some reviews reports about using drugs recommended by relatives (207 reviews), advertisement (97) or internet (15).

\begin{figure*}
    \centering
    \includegraphics[width=12cm]{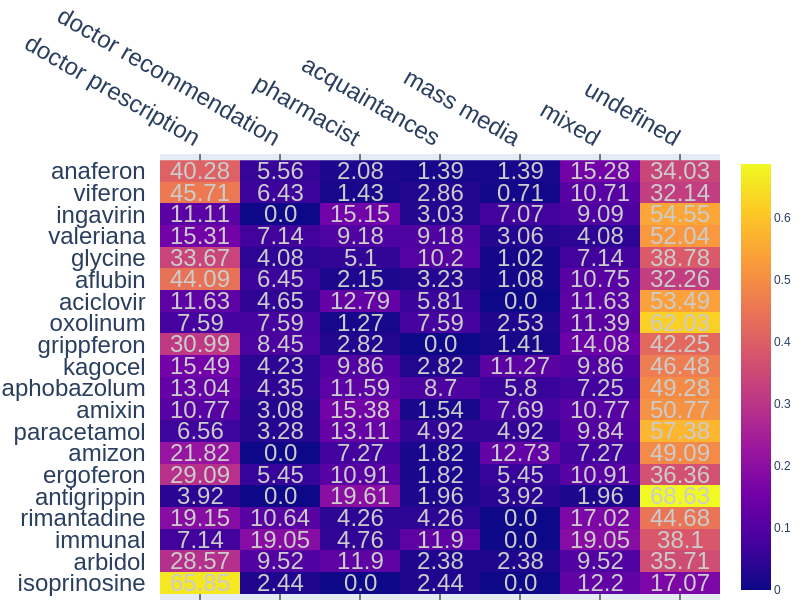}
    \caption{The distribution heatmap of reviews percentages for different sources of information for the 20 most popular drugs. The number in a cell means the percentage of reviews with the drug and particular source to the total number of reviews with this drug. If there were several different sources mentioned, it counted as ``mixed'' source}
    \label{fig:top15_drug_heatmap}
\end{figure*}

The heatmap, presented on Fig.~\ref{fig:top15_drug_heatmap}, shows percentages of reviews where popular drugs were co-occurred with different sources (sources were manually merged into 5 groups by annotators).
It could be seen that most recommendations are coming from professionals. For example Isoprinosine (used in 65.85\% cases by medical prescription), Aflubun (44.09\%), Anaferon (47.30\%) and others. However, for such drugs as Immunal (11.9\%) or Valeriana (9.18\%) the rate of usage on the advice of patients' acquaintances is close to doctors' recommendations or higher. Amizon (12.73\%) and Kagocel (11.27\%) have the highest percentage for mass media (advertisement, internet and other) as the source  compared to other drugs.

\begin{figure}
    \centering
    \includegraphics[width=8.5cm]{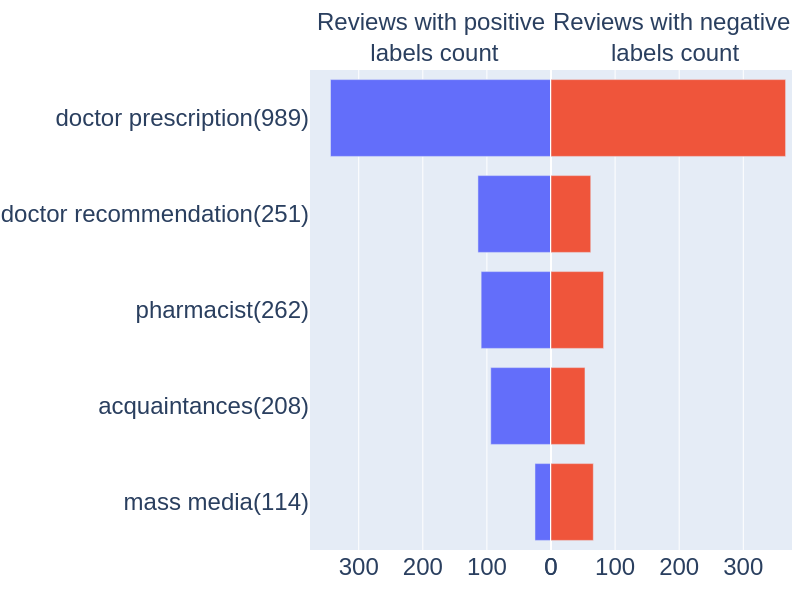}
    
    \caption{Distribution of the tonality for the different sources. Number in brackets shows reviews count with the source of information, including reviews without reported effects or neutral reviews (with both good and bad effects)}
    \label{fig:Distribution_drug_tonality}
\end{figure}

The distribution of the tonality (positive or negative) for the sources of information is presented in Fig.~\ref{fig:Distribution_drug_tonality}. A source is marked as ``positive'' if positive dynamic is appeared after the use of drug (i.e. review includes ``BNE-pos'' attribute). ``Negative'' tonality is marked if negative dynamic or deterioration in health has taken place or drug has had no effect (i.e. ``Worse'', ``ADE-Neg'' or ``NegatedADE'' mentions appear). Reviews with both effects were not taken into account. It follows from the diagram that drugs recommended by doctors or pharmacists are mentioned more often as having positive effect, while using drugs based on an advertisement often leads to deterioration in health.

\begin{figure*}
    \centering
    \includegraphics[width=17cm]{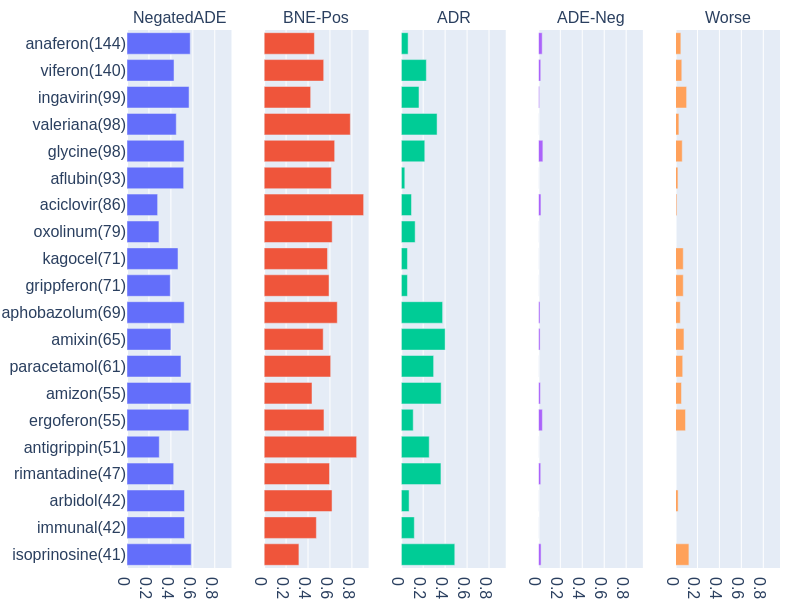}
    \caption{Distributions of labels of effects reported by reviewers after using drugs. Top 20 drugs by the reviews count are presented. The number in brackets is the number of reviews with mentions of a drug. Diagrams show part of reviews mentioning a specific type of effect from the total amount of reviews with the drug}
    \label{fig:effects}
\end{figure*}

Diagrams in Fig.~\ref{fig:effects} show parts of reviews where popular drugs were mentioned along with labeled effects. The following drugs have largest parts for ADR in reviews: immunomodulator -- ``Isoprinosine'' (48.8\% of reviews with this drug contains mentions of ADR), antiviral ``Amixin'' (40.0\%),  tranquilizer -- ``Aphobazolum'' (37.7\%), antiviral -- ``Amizon'' (36.4\%), antiviral -- ``Rimantadine'' (36.3\%).

Users mention that some drugs causing negative dynamics after start or some period of using it (ADE-Neg). Examples of such drugs are ``Anaferon'' (3.5\% of reviews with this drug mention ADE-Neg effects), ``Viferon'' (2.1\%), ``Glycine'' (4.1\%), ``Ergoferon'' (3.6\%). 

According to reviews some of the drugs causes deterioration in health after taking the course (``Worse'' label): immunomodulator -- ``Isoprinosine'' (12.2\%), antiviral -- ``Ingavirin'' (10.1\%), ``Ergoferon'' (9.1\%) and other.


This corpus is used further to get a baseline accuracy estimate for the named entity recognition task.

\subsection{Coreference annotation}\label{subsec:Coreference_annotation}
To begin with, we used a state-of-the-art neural network model for coreference resolution~\cite{joshi2019bert}, and adapted it to Russian language by training on the corpus AnCor-2019.  After this we predicted coreference for reviews in our corpus. We chose 91 reviews which had more that 2 different drug names and disease names (after manual grouping described in \ref{subsec:Normalization})and more than 4 coreference clusters and 209 reviews which had more that 2 different drug names and more that 2 coreference clusters. These 300 reviews we gave to our annotators for manual checking of coreference clusters, predicted by model.

The annotators had guideline for coreference and a set of examples. According to guidelines they supposed to pay attention to mentions annotated with pharmacological types, pronouns and words typical for references (e.g. ``such'', ``former'', ``latter''). They didn't annotate as coreference following things:
\begin{itemize}
    \item mentions of reader (``I wouldn't recommend you to buy it if you don't want to waste money'');
    \item  split antecedents - when 2 or more mentioned entities also mentioned by a common phrase (``I tried Coldrex and after a while i decided to buy Antigrippin. Both drugs usually help me.'');
    \item generic mentions - phrases that describe some objects or events(e.g. ``Many doctors recommend this medication. Since I respect the opinion of doctors I decided to buy it.'' - doctors are not coreferent mentions);
    \item phrases that gives definitions to other (``Valeriana is a good sedative drug that usually helps me'' - ``Valeriana'' and ``sedative drug'' are not coreferent mentions).
\end{itemize}

The table \ref{tab:coref_number_comparison} shows the number of coreference clusters and mentions in 300 drug reviews from our corpus compared to corpus AnCor-2019. It should be noted that not all coreference mentions correspond to mentions of our main entity annotation, sometimes a single coreference mention can unite multiple medical mentions,or connect pronouns that are not involved in medical annotations.  The table \ref{tab:coref_medtypes} represents the number of medical mentions of various types that intersect coreference mentions. This corpus is used further to get a baseline accuracy estimate for the named entity recognition task.

\begin{table}
\centering
\caption{Number of coreference chains and mentions compared to other Russian coreference corpus} \label{tab:coref_number_comparison}
\input{tables/coref_number_comparison.tex}
\end{table}

\begin{table}
\centering
\caption{Mentions types involved in coreference chains} \label{tab:coref_medtypes}
\input{tables/coref_medtypes.tex}

\end{table}

\section{Machine learning methods}\label{sec:Methods} 
\subsection{Entities detection problem}\label{subsubsec:Methodology} 

    \label{Model}
We consider the problem of named entity recognition as a multi-label classification of tokens -- words and punctuation marks -- in sentences. 
  Phrases of different entities can intersect, so that one word can have several tags. 

The output for each token is a tag in the BIO format: the ``B'' tag indicates the first word of a phrase of the сonsidered entity, the ``I'' tag is used for subsequent words within the mention, and the ``O'' tag means that the word is outside of an entity mention.



To set the accuracy level of entity recognition in our corpora we used two methods. The first (Model A) was based on BiLSTM neuralnet topology with different feature representation of input text: dictionaries, part of speech tags and several methods of word level representations, incl. FastText~\cite{bojanowski2017enriching}, ELMo~\cite{peters2018deep}, BERT, words character LSTM coding, etc. 
The second (Model B) was a multi-model combining the  pretrained multilingual language model XLM-RoBERTa~\cite{xlm_conneau2019unsupervised} and the LSTM neural network  with several most effective features. 
Details of the implementation of both methods with a description of the used features are presented below. 

\subsection{Used features}\label{subsubsec:features} 
\paragraph{Tokenization and Part-of-Speech tagging.} 
To preprocess the text we used UDPipe~\cite{straka2016udpipe} tool. After parsing each word get 1 of 17 different parts of speech. They are represented as a one-hot vector and used as an input for the neural network model. For model B, the text was segregated on phrases using UDPipe version 2.5. Long phrases splitted up into 45 word chunks.

\paragraph{Common features.} They are represented as a binary vector of answers to the following questions (1 if yes, 0 otherwise):
\begin{itemize}
    \item Are all letters capital?
    \item Are all letters in lowercase?
    \item Is the first letter capital?
    \item Are there any numbers in the word?
    \item Does more than a half of the word consist of numbers?
    \item Does the entire word consist of numbers?
    \item Are all letters Latin?
\end{itemize}


\paragraph{Emotion markers.} Adding the frequencies of emotional words as extra features is motivated by the positive influence of these features on determining the author's gender~\cite{SueroMontero201498}.
Emotional words are taken from the dictionary
~\cite{emofeelDicts} 
which contains 37 emotion categories, such as <<Anxiety>>, <<Inspiration>>, <<Faith>>, <<Attraction>>, etc.
On the basis of the $n$ available dictionaries, an $n$-dimensional binary vector is formed for each word, where each vector component reflects the presence of the word in a certain dictionary.

In addition, this word feature vector is concatenated with emotional features of the whole text. These features are LIWC and psycholinguistic markers.

The former is a set of specialized English Linguistic Inquiry and Word Count (LIWC) dictionaries~\cite{tausczik2010psychological}, adapted for the Russian language by linguists~\cite{litvinova2017deception}. The LIWC values are calculated for each document based on the occurrence of words in specialized psychosocial dictionaries. 

Psycholinguistic text markers~\cite{sboev2015quantitative} reflect the level of the emotional intensity of the text. They are calculated as the ratio of certain frequencies of parts of speech in the text. We use the following markers: the ratio of the number of verbs to the number of adjectives per unit of text; the ratio of the number of verbs to the number of nouns per unit of text; the ratio of the number of verbs and verb forms (participles and adverbs) to the total number of all words; the number of question marks, exclamation points, and average sentence length. The combination of these features are referred to as "ton" in Table~\ref{tab:feat_top_exp}.

\paragraph{Dictionaries.} 
The following dictionaries from open databases and registers are used as additional features for the neural network model.
\begin{enumerate}
    \item Word vectors formed on base of the MESHRUS thesaurus as described in Appendix~\ref{subsec:meshrus_features}. The two approaches described in that section are referred to as MESHRUS and MESHRUS-2. The resulting CUI codes are encoded with one-hot representation.
    \item \textbf{Vidal}. 
    For each word, a binary vector is formed, which reflects belonging to categories from the Vidal medication handbook~\cite{Vidal2019}: adverse effects, drug names in English and Russian, diseases. The dataset words are mapped to the words or phrases from the Vidal handbook. To establish the categories, the same approach as for MESHRUS is used. The difference is that instead of setting indices for every word (as CUI in the UMLS) we assign a single index to all words of the same category. That way, words from the dataset are not mapped to special terms, but checked for category relations.
\end{enumerate}

\begin{figure}
\centering
\includegraphics[scale=0.37]{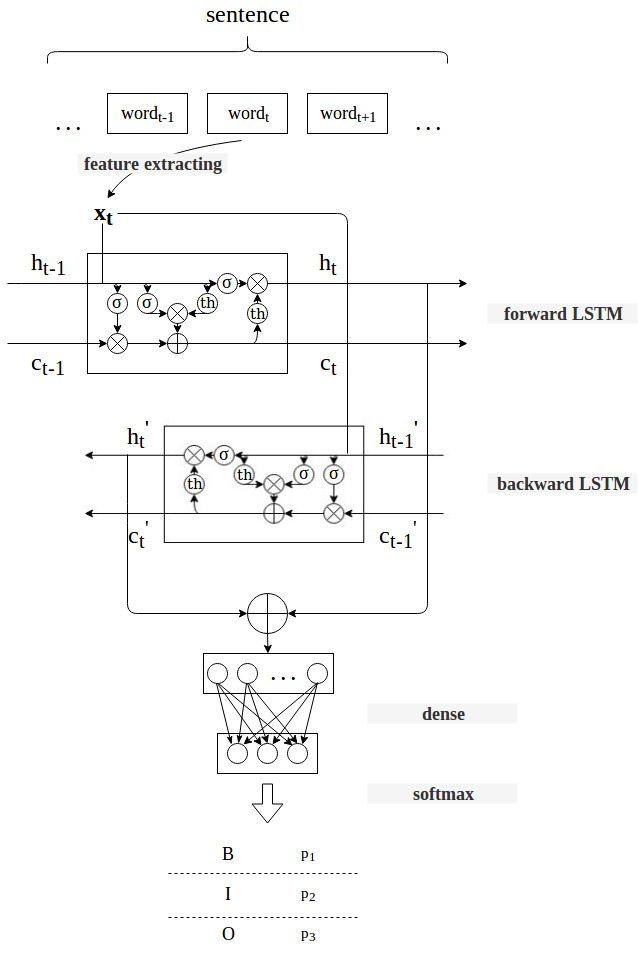}
\caption{The main architecture of the network. Input data goes to bidirectional LSTM, where the hidden states of forward LSTM and backward LSTM get concatenated, and the resulting vector goes to fully-connected layer with size 3 and SoftMax activation function. The output $p_{1}$, $p_{2}$, and $p_{3}$ are the probabilities for the word to belong to the classes B, I, and O, i.\,e. to have B, I, or O tag.}
\label{fig:lstm}
\end{figure}

\subsection{Word vector representations} 
It is the representation of word by a vector in a special space where words with similar meanings are close to each other. 
The following models were used: FastText~\cite{bojanowski2017enriching}, ELMo (Embeddings from Language Model)~\cite{peters2018deep}, and BERT (Bidirectional Encoder Representations from Transformer)~\cite{devlin2018bert}, XLM-RoBERTa~\cite{xlm_conneau2019unsupervised}.
The approach of the FastText is based on the Word2Vec model principles: word distributions are predicted by their context, but FastText uses character trigrams as a basic vector representation. Each word is represented as a sum of trigram vectors that are the base for continuous bag of words or skip-grams algorithms~\cite{mikolov2013efficient}. Such a model is simpler to train due to decreased dictionary size: the number of character n-grams is less than the number of unique words. Another advantage of this approach is that morphology is accounted automatically, which is important for the Russian language.

Instead of using fixed vectors for every word (like FastText does), ELMo word vectors are sentence-dependent. ELMo is based on The Bidirectional Language Model (BiLM), which learns to predict the next word in a word sequence. Vectors obtained with ELMo are contextualized by means of grouping the hidden states (and initial embedding) in a certain way (concatenation followed by weighed summation).
However, predicting the next word in a sequence is a directional approach and therefore is limited in taking context into account. This is a common problem in training NLP models, and is addressed in BERT.

BERT is based on the Transformer mechanism, which analyzes contextual relations between words in a text. The BERT model consists of an encoder extracting information from a text and a decoder which gives output predictions. In order to address the context accounting problem, BERT uses two learning strategies: words masking and logic check of the next sentence. The first strategy implies replacing 15\% of the words on a token ``MASK'' which is later used as a target for the neural network to predict actual words. In the second learning strategy, the neural network should determine if two input sentences are logically sequenced or are just a set of random phrases. In BERT training, both strategies used simultaneously so as to minimize their combined loss function.
XLM-RoBERTa model model a similar to BERT masked language model based on Transformers~\cite{vaswani2017attention}. Main differences between XLM-RoBERTa and BERT are following: XLM-RoBERTa was trained on larger multilingual corpus from CommonCrawl project which contains 2.5TB of texts. Russian is the second language by texts count in this corpus after English. XLM-RoBERTa was trained only for masked token prediction task, it didn't use the next sentence prediction loss. Minibatches during model training included texts in different languages. It used different tokenization algorithm, while BERT used WordPiece~\cite{schuster2012japanese}, this model used SentencePiece~\cite{kudo2018sentencepiece}. Vocabulary size in XLM-RoBERTa is 250K unique tokens for all languages. There is two versions of model: XLM-RoBERTa-base with 270M parameters and XLM-RoBERTa-large with 550M.

\subsection{Model architecture}
\subsubsection{Model A - BiLSTM neural net}
\begin{figure}
    \centering
    \includegraphics[scale=0.3]{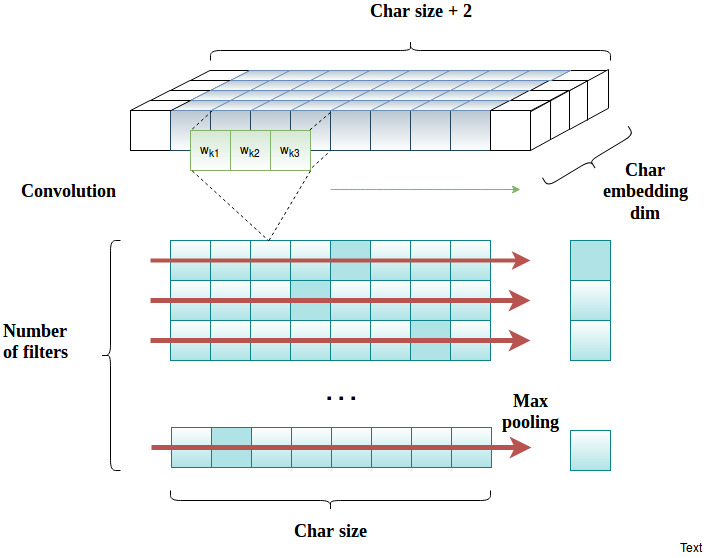}
    \caption{The scheme of character feature extraction on base of char convolution neural network. Each input vector after the embedding layer is expanded with two extra padding object (white boxes), $w_{(k1)}, w_{(k2)}, w_{(k3)}$ - weights of convolution filter $k$.}
    \label{fig:char_cnn}
\end{figure}

The topology of Model A is depicted in Fig. ~\ref{fig:lstm}. The set of input features for this model was described above.
Additionally for word coding we used characters convolution based neural network (see Fig. ~\ref{fig:char_cnn}), CharCNN~\cite{krizhevsky2012imagenet}.
First, each word is represented as a character sequence. The number of characters is a hyperparameter, which in this study has chosen empirically with the value of 52. If the word has fewer characters than this number, the remaining characters are filled with the <<PADDING>> symbol. The training dataset is used to make a character vocabulary that also includes special characters <<PADDING>> and <<UNKNOWN>>, the latter allowing for possible future occurrence of characters not present in the training set.
For coding each character embedding layer~\cite{gal2016theoretically} is used, which replaces every character from vocabulary appeared in a word to a corresponding real vector. In the beginning, the real vectors are initialized with values from random uniform distribution in the range of [-0.5; 0.5]. The size of real vectors is 30. Further, the matrix of coded characters of word is processed by convolution layer (with 30 filters and kernel size = 3)~\cite{dumoulin2016guide} and global maxpooling function that provided maximization function of all values for each filter~\cite{boureau2010theoretical}.

At the output of the model, we put either a fully connected layer~\cite{chiu2016named} or conditional random fields (CRF~\cite{lafferty2001conditional}), which output the probabilities for a token to have a B, I, or O tag for the corresponding entity (for instance, B-ADR, I-ADR, or O-ADR).

\subsubsection{Model B - XLM RoBERTa based multi-model}

\begin{figure}
    \centering
    \includegraphics[width=8.0cm]{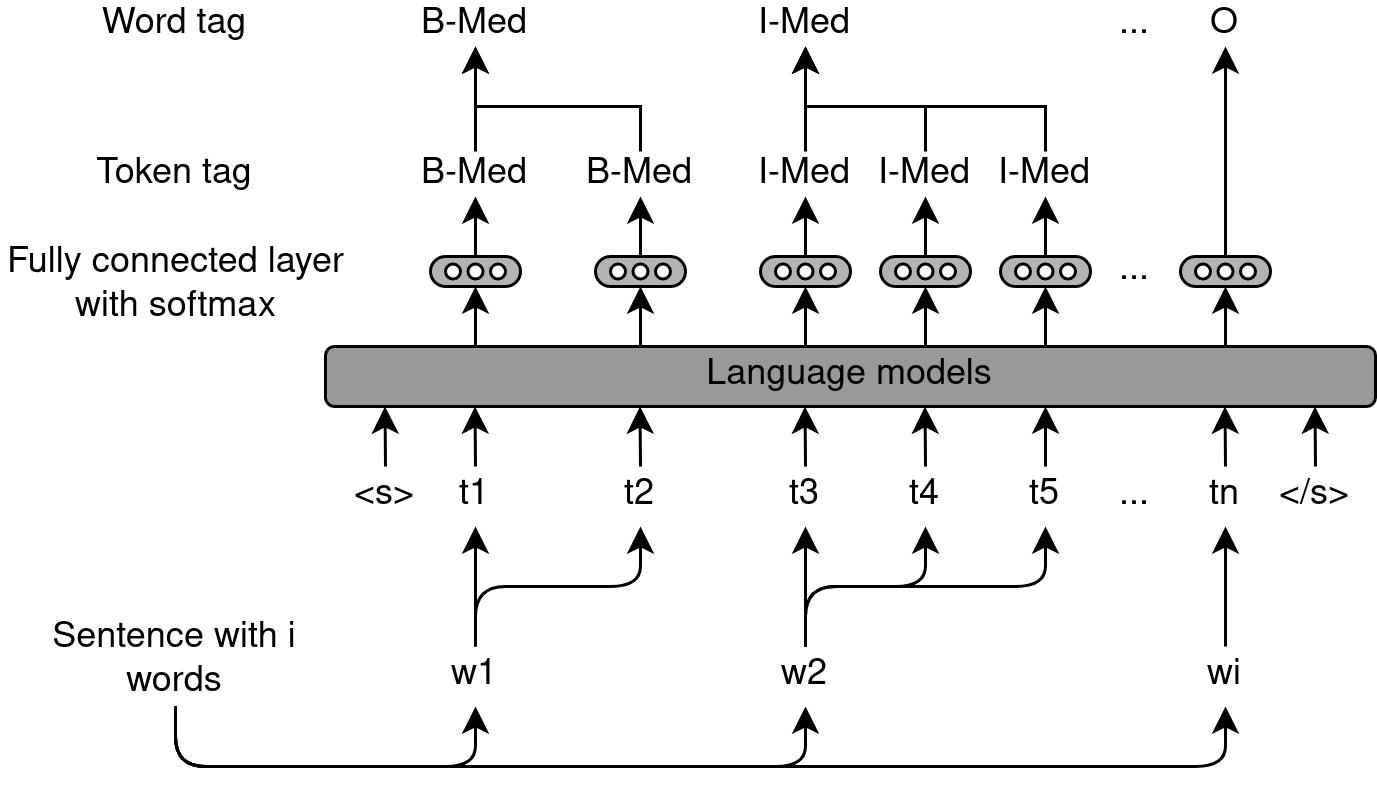}
    \caption{Fine tuning of language model for word classification task}
    \label{fig:ner_finetune}
\end{figure}

\begin{figure}
    \centering
    \includegraphics[width=8.0cm]{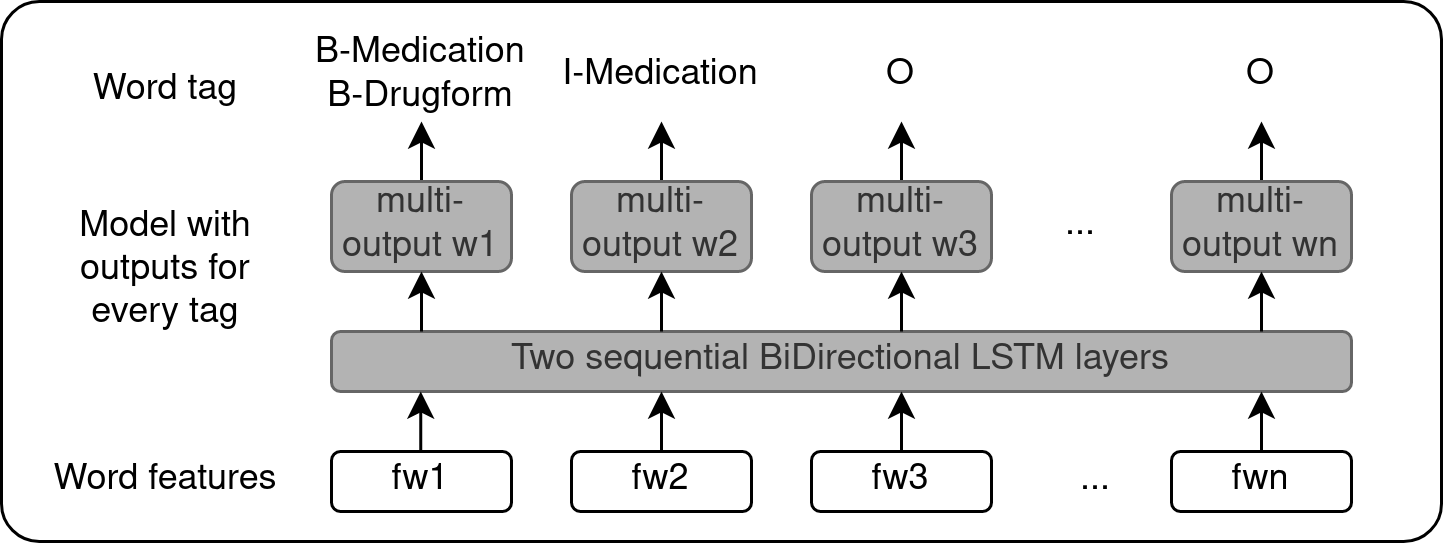}
    \caption{Model B architecture, fwn -- words, encoded with features}
    \label{fig:modelB_output}
\end{figure}

\begin{figure}
    \centering
    \includegraphics[width=8.0cm]{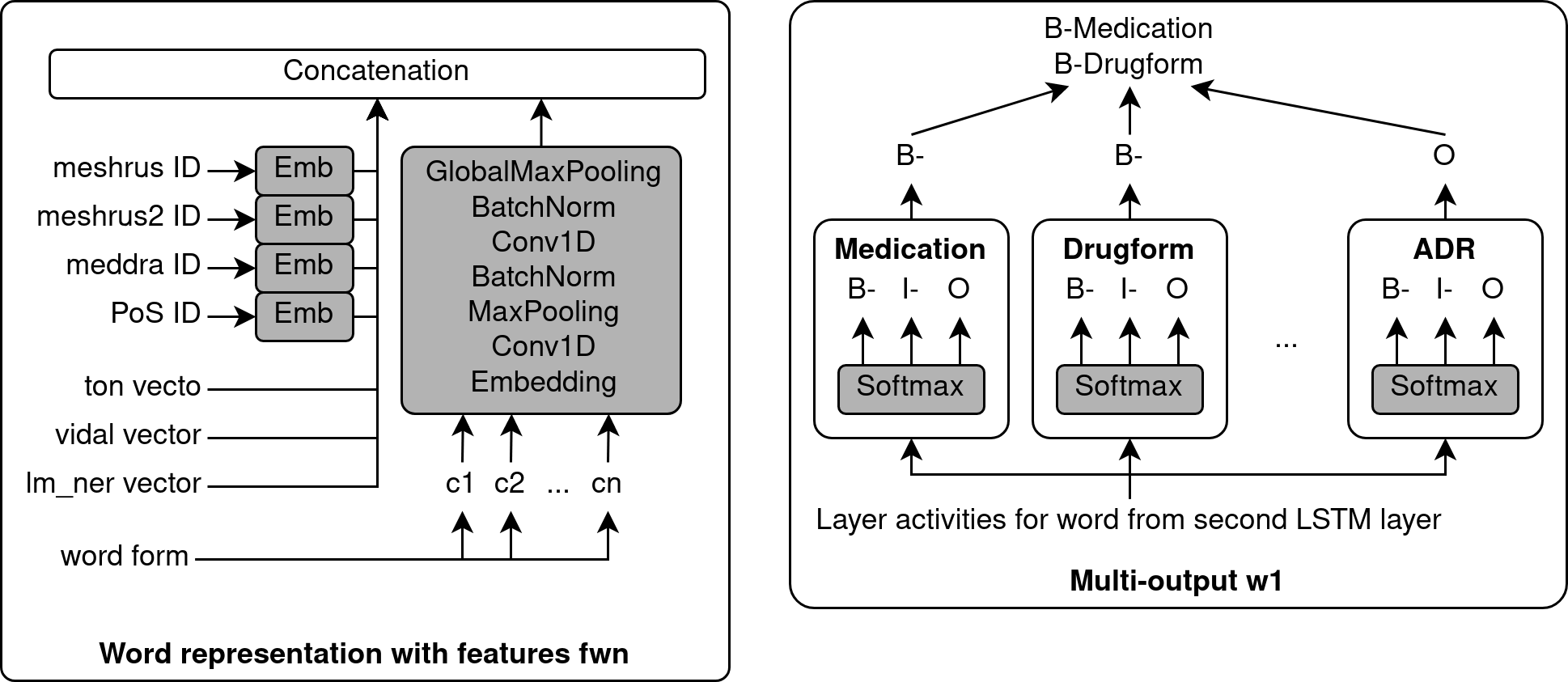}
    \caption{Model B architecture, words encoding method to obtain word representation vectors fwn is presented on the left and on the right there is multi-output for words classification}
    \label{fig:modelB_encoding}
\end{figure}

To improve the model accuracy, we performed an additional training XLM-RoBERTa-base on two datasets:  the first we collected from the site \url{irecommend.ru} and the second was borrowed from unnannotated part of RuDReC \cite{tutubalina2020russian}.  Calculations of two epochs during three days and XLM-RoBERTa-large for one epoch during 5 days were performed using a computer with one Nvidia Tesla v100 and Huggingface Transformers library.  Further, we fine-tuned these models to solve the NER task. Figure \ref{fig:ner_finetune} demonstrates an algorithm of fine-tuning language models for NER. This is the commonly used fine-tuning algorithm of simple transformers project~\cite{rajapakse2019simpletransformers}. The linear layer with an activation function softmax was added to the model output to classify words. The developed multi-tag model implements the concatenation of fine-tuned language model with the vector of features (Vidal, MESHRUS, ton, and other). The LSTM neural net model processes then the resulting vector to implement the multi-tagged labeling. Figures \ref{fig:modelB_output}, \ref{fig:modelB_encoding} clarify a model topology. So the multi-tag model combines the above-mentioned fine-tuned language model with the simplified variant of Model A(without CRF and with the substitution of ELMo  word representation by the fine-tuned language model's output with class activities). During training the above-mentioned LSTM neural net model, this language model was not trained. We used the automatic selection of hyperparameters using Weights\&Biases~\cite{wandb} – sweeps for the total multi-tag model. It took about 24 hours on the computer with 3 Tesla K80 processing 6 agents. The 5-fold evaluation was used.

\subsubsection{Coreference model}
For coreference resolution, we chose a state-of-the-art neural network architecture from \cite{joshi2019bert}.  The core feature of this model is the ability to learn the task of mentions detection, and the task of mentions linking and forming coreference clusters end to end at the same time, without separating these 2 tasks into different processes. The model uses the BERT language model to get input text word vector representations.
To adapt network architecture to Russian language we used RuBERT - BERT language model trained on the Russian part of Wikipedia and news data. We conducted experiments to tune neural network hyperparameters and training options to achive a better results, final hyperparameters were as follows: maximum span width = 30, maximum antecedents for every mention: 50, hidden fully connected layers size = 150, numbers of sequential hidden layers = 2,maximum epoch training: 200, language model learning rate = 1.0e-05, task model learning rate = 0.001,embedding sizes = 20.
  
\section{Experiments}\label{subsec:Experiments}
\subsection{Methodology}
In the experiments, we pursued the following objectives: 
\begin{enumerate}
    \item To select most effective language model among the set: FastText, ELMo, and BERT;
    \item To evaluate the influence of different feature sets on the precision of  ADR mention extraction;
    \item To compare the level of precision for ADR mentions identification basing on our corpus in relation to one received  on available russian language data of similar type;
    \item To show  the influence of  such characteristics of  corpus texts on the precision of  ADR mention extraction, as the proportion between phrases with ADR  and without  it, between ADR mentions and INDICATION mentions, the corpus size and etc.         \item To evaluate the influence of the ADR tagging severity on the ADR identification precision.
\end{enumerate}

We made the accent on ADR because of its importance in practice and the complexity of identification given close relation to the context that stipulates this selection for model calibrations.
\begin{table*}
\caption{Accuracy (\%) of recognizing ADR, Medication and Disease entities in our corpus (1600 reviews) by Model A with different language models.}
\label{tab:embedding}
\centering
\input{tables/embedding_exp.tex}
\end{table*}

\label{section:quality_metrics}
For models performance estimation, we used the chunking metric, which was introduced in the conll2000 shared task and has been used to compare named entity extraction systems since then. The implementation can be found here: \url{https://www.clips.uantwerpen.be/conll2000/chunking/}. The script receives as its input a file where each line contains a token, true tag and predicted tag. Tags could be "O" - if token doesn't belong to any mentions, "B-X" if token starts a mention of some type X, "I-X" if it continue a mention of type X. If tag "I-X" appears after "O", or "I-Y" (mention of other type) it's treated as "B-X" and starts a new mention. The script calculates the percentage of detected mentions that are correct (precision), the percentage of correct mentions that were detected (recall) and an $F_1$ score: $$F_1 = \frac{2*\mathrm{precision}*\mathrm{recall}}{\mathrm{precision} + \mathrm{recall}}$$
In our work we use F1-exact score that estimate accuracy of full entity matching. 




\label{section:pipeline}
   \subsection{Finding the best embedding}

    We considered the following embedding models: FastText, ELMo, and BERT. Two corpora were used to train the FastText model -- a corpus of reviews from Otzovik.com from the category "medicines" and a corpus of reviews from the category "hospitals"~\footnote{Reviews were taken from the Otzovik website from the categories "hospitals" and "medicines" - https://otzovik.com/health/},
    also we used vectors pretrained on the Commoncrawl corpus\footnote{\url{http://commoncrawl.org/}}.
    The ELMo model which had been preliminarily trained on the Russian WMT News~\cite{statmt} was taken from the DeepPavlov~\footnote{\url{https://deeppavlov.readthedocs.io/en/master/intro/pretrained_vectors.html}}~\cite{burtsev2018deeppavlov} open-source library. The pretrained multilingual BERT model was taken from the Google repository~\footnote{\url{https://github.com/google-research/bert/}} and subsequently fine-tuned on the above-mentioned corpora of drug and hospital reviews. These pretrained models were used as input to our neural network model presented in Fig.~\ref{fig:lstm}. The dataset (the first version of our corpus contained 1600 reviews) was split into 5 folds for cross-validation. On each fold, the training set was split into training and validation sets in the ratio 9:1. Training was performed for a maximum of 70 epochs, with early stopping by the validation loss. Cross entropy was used as the loss function, with nAdam as the optimizer and cyclical learning rate mechanism~\cite{smith2017cyclical}. The results of the test experiments are given in Table~\ref{tab:embedding}, where the best results according to the F1-exact metric demonstrate ELMo. The composition of ELMo with BERT    worsens the precision. As a result, we used ELMo below to evaluate the influence of different features on ADR mention extraction precision.
    
    
    
    
    \subsubsection{The influence of different features on ADR recognition precision} To evaluate the influence of using any separated feature from those mentioned above on ADR precision, we conducted the series of experiments with Model A which results presented in Table \ref{tab:feat_top_exp}.
    \subsubsection{Choosing the best model topology}
    Next, we provide a set of experiments with Model A on the choice of topology: replacing the last fully-connected layer with a CRF layer, or changing the number of biLSTM layers. This was studied in combination with adding emotion markers, PoS and MESHRUS, MESHRUS-2 and Vidal dictionaries, as shown in Table~\ref{tab:feat_top_exp}. So, this made it possible to assess the accuracy level of Model A. To evaluate the effectiveness of XLM-RoBERTa-large, we ran it without features (see last row in Table~\ref{tab:feat_top_exp}). In view of the it's high precision exceeding the precision of Model A, we used it as basis to  create Model B.    

\begin{table*}
\centering
\caption{Entity recognition F1 score on our corpus (1600 reviews) of the models with different features and topology.}
\label{tab:feat_top_exp}

\input{tables/feature_top_exp.tex}
\end{table*}

 \subsubsection{The influence of  characteristics of corpus texts on the precision of ADR recognition} 
 First of all, we conducted experiments on the corpus 2800 texts extended by texts similar to corpus 1600 texts to assess the change of precision in ADR identification with the rise of ADR mention number. As follows from the data in Table~\ref{tab:rdrs_subsets}, a direct increase in the number of reviews in the corpus gives only a small increase in the share of ADR-mentions per review (0.2 versus 0.22). So, its saturation by ADR stays lower than in most corpora from Table~\ref{tab:adr_saturation}. To study the effect of increasing saturation of the corpus by ADR mentions, we experimented with  sets of different sizes from the corpus with various ADR-mention shares per review:   of 1250 texts (average 1.4 ADR onto review) balanced with ADR and without ADR, of 610 texts(average 2.9 ADR onto review), of 1136 texts(average 1.5 ADR onto review), of 500 texts (average 1.4 ADR onto review). In all experiments, the model treated input texts as the set of independent phrases.  
  \subsubsection{The influence evaluations of annotation style of ADR on its recognition precision}
  In this case, we ran two experiments to evaluate a difference in  ADR mention extractions: the first on the base of the set containing pure  ADR mentions and the second, including the doubtful bordering ADR mentions, annotated both ADR and NOTE.
  
\begin{table*}
\centering
\caption{Subsets of RDRS corpora with accordance to complexity of ADR level saturation. *Model B - XLM-RoBERTa part only score on RuDRec}

\input{tables/rdrs_subsets.tex}

\label{tab:rdrs_subsets}
\end{table*}

 \subsubsection{Evaluations of the precision of coreference  relations extractions on our corpus by models  trained on different corpora} After  annotators  manually  corrected  predicted coreference relations in our corpus, we splited it to train, validation and test subsets. Then we evaluated coreference resolution model trained on AnCor-2019 corpus and tested on our corpus and model trained on our corpus.  We also did same experiments on AnCor-2019 test subset. We also tried to combine both train sets.
  

\section{Results}\label{sec:results}

\subsection{Results of Model A in series of embedding comparison experiments} These results are presented in Table~\ref{tab:embedding} and demonstrate the superiority of the ELMo model. BERT leads to lower F1 values with larger deviation ranges, and with the FastText model the F1 score is the lowest. Consequently, in further experiments on adding features and changing the topology we use the ELMo embedding as the basic approach. 
The composition of ELMo with BERT worsens the precision. As a result, we used ELMo below to evaluate the influence of different features on ADR mention extraction precision.
    
\subsection{Results of choosing the best model topology and input feature set for Model A in comparison with XLM-RoBERTa-large results}
For our corpus, as shown in Table~\ref{tab:feat_top_exp}, various changes in features and topology were added to the basic model with ELMo embedding. First of all, we focused on the metric $\Fe$, since it reflects the quality of the model better. Adding features gave the greatest increase in the least-represented class ADR. As a result, a combination of dictionary features, emotion markers, 3-layers LSTM and CRF can achieve the highest quality increase in ADR and Disease entities. For Medication, the combination of ELMo and 3-layer LSTM showed slightly better results. But results of experiments with model A as a whole are worse than the results of XLM-RoBERTa-large, which was used as a basis of Model B. Therefore, we performed further experiments on the base of Model B founded on it, as the best.

\subsection{Results  of the influence evaluation of corpus texts characteristics on the precision of ADR mention recognition }

\begin{figure*}
    \centering

    \includegraphics[width=\textwidth]{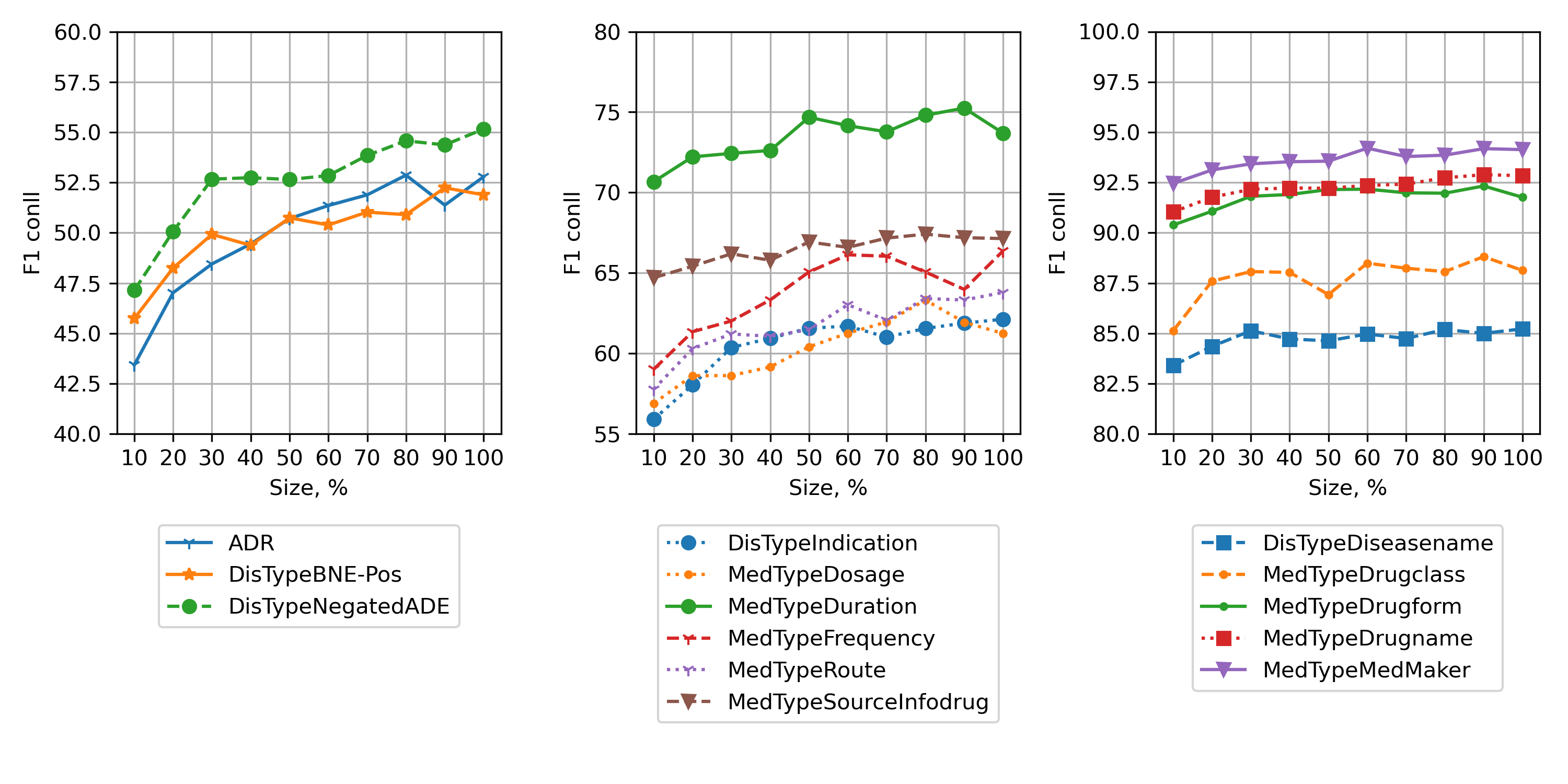}
    
    \caption{Dependency of the accuracy on the size of training set for different tags in RDRS 2800}
    \label{fig:size_and_f1_g1}
\end{figure*}

The direct rise of corpus volume up from 1600 to 2800 mentions results in the ADR identification precision increase on 13\% F1, 6\% F1 in Disease, 4\% F1 in Medication. Figure \ref{fig:size_and_f1_g1} shows a curve of dependence of ADR precision increase on the corpus size, which becomes stable out of 80\% corpus size. Such behaviour for other main subtags demonstrate similar courses (see Table \ref{tab:all_tags_model_B}). The rise of the  ADR share by balancing the corpus leads to a more significant increase in ADR precision on 21\% without significant Disease and Medication precision identification changes (see Table ~\ref{tab:integrate_scores}). The higher saturation by these tags, which in practice stays unchanged after balancing corpus, explains the last fact. Experiments on corpora with the saturation more closer to CADEC one showed the further increase of  the ADR identification precision, up to 71.3\% F1 on the corpus of 610 texts ADR (average 2.9 ADR onto review).

\begin{table}
\centering
\caption{F1-scores of the model B for RDRS 1250 and RDRS 2800. *Negative -- union of tags: Worse, NegatedADE, ADE-Neg.}
\input{tables/all_tags_2800_model_B.tex}
\label{tab:all_tags_model_B}
\end{table}

\begin{table}
\centering
\caption{The difference in accuracy for the 3 main tags depending on the size and balance of the corpus} \label{tab:integrate_scores}
\input{tables/integral_scores.tex}
\end{table}

\subsection{Results of experiments to evaluate the influence of annotation style of ADR mentions on ADR recognition precision}
This case results of experiments on the balanced set allowed evaluating the effect of the relaxation of ADR annotation requirements in about 3\% of the precision increase as follows figure (see Fig.~\ref{fig:adr_saturation}).
\subsection{Results for the coreference model} Results, presented in table 14, shows that  the used model trained on the  subset of our corpus demonstrates a high result  on the   test subset of our corpus. The training on AnCor-2019 corpus or on  corpora AnCor-2019  with ours gives worse results.

\begin{table*}
\centering
\caption{Results of training coreference resolution model on different corpora}
\label{tab:coref_eval}
\input{tables/coref_eval.tex}
\end{table*}


\section{Discussion}
\label{sec:discussion}
Currently, there are a significant diversity of full-sized labeled corpora in different languages to analyze the safety and effectiveness of drugs.  We present the first full-size Russian compound NER-labeled corpus - RDRS - of Internet user reviews with the  labelled coreference relations  in part of the corpus. Based on the developed neural net models results, we investigated this corpus place in this diversity depending on the corpora characteristics. The analysis made on base of the experiment sequence sets of different  saturation  by  the definite entity  extracted from the corpus allows to give the more realistic conclusion  about its quality in concerns to this entity. The results of developed model B on base of   XLM-RoBERTa-large outperforms the results of work~\cite{tutubalina2020russian} on 2.3\% for ADR recieved on the corpus of limited size that grounds  a quality of developed model B and an applicability of its results to  establish  the state of the art precision level of entity extraction on the created corpus. 

\begin{figure}
    \centering

    \includegraphics[width=8.5cm]{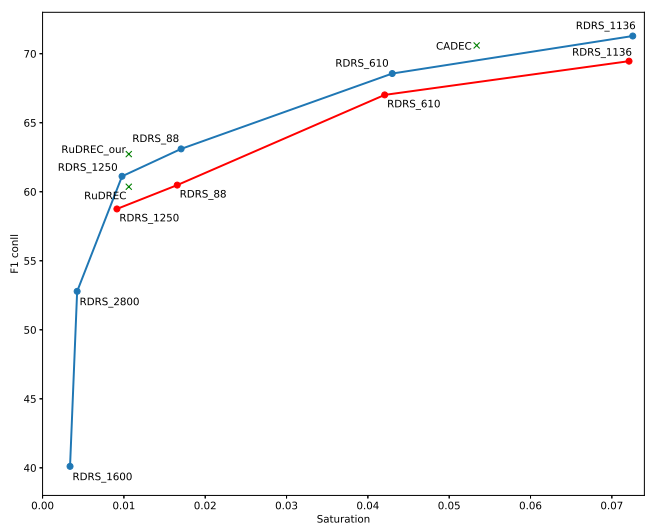}
    
    \caption{Dependency of ADR recognition precision on their saturation in the corpora. Red line  - different subsets of our corpus (see Table~\ref{tab:rdrs_subsets}) with pure ADR annotation. Blue line - different subsets of our corpus with doubtful bordering annotation (annotated both ADR and NOTE), RuDREC - published accuracy for RuDREC corpus~\cite{tutubalina2020russian}, RuDREC\_our - our accuracy for RuDREC corpus, CADEC - published accuracy for CADEC corpus~\cite{li2020lexicon}}. 
    \label{fig:adr_saturation}
\end{figure}

In general, the results of experiments with sets of different sizes and different saturation showed that  in case of ADR mention, a strong dependence  of the  ADR identification precision on corpus saturation  by them  exists (see Figure~\ref{fig:adr_saturation}). So the comparison of our corpus with any one of the close types, such as the СADEC, is necessary to conduct on the dataset of corpus examples with the close saturation by ADRs. The coreference relation extraction experiments show that despite the AnChor-2019 corpus is greater in the number of relations than our corpus; both corpora demonstrate similar precisions when the training set is from the first corpus, the testing set is from the other. But in the case of training on the set of examples from both corpora, we received worse results—these results directly on the essential difference in compositions of the corpora from different domains.

\section{Conclusion}\label{sec:conclusion}
The primary basic result of this work is the creation of the Russian full-size  NER multi-tag labeled corpus of the Internet user reviews, including the part of the corpus with annotated coreference relations. The multi-labeling model appropriated for presented corpus labeling based on combining a language model XLM-RoBERTa with the selected set of features is developed. The results obtained basing this model showed that the accuracy level of ADR extraction on our corpus is comparable to that obtained on corpora of other languages with similar characteristics. Thus, this level may be seen as state of the art on this task decision on Russian texts in view. The presence of the corpus part with annotated coreference relations allowed us to evaluate the precision of their extraction on texts of the profile under consideration.

 The developed neuronet complex may be used as a base for the replenishment of the corpus by ADR. This, along with including new entities and relations, is a goal of further work.

\section*{Acknowledgments}
This work has been supported by the Russian Science Foundation grant 20-11-20246 and carried out using computing resources of the federal collective usage center Complex for Simulation and Data Processing for Mega-science Facilities at NRC “Kurchatov Institute”, http://ckp.nrcki.ru/.

\bibliographystyle{cas-model2-names}

\bibliography{cas-dc-template}

\appendix
\section{Appendix}
\subsection*{ADR recognition on the basis of the PsyTAR corpus} \label{subsec:psytar_adr}
PsyTAR corpus from  \cite{basaldella2019bioreddit} contains sentences in a CoNLL format.
This modification of a corpus is publicly available \footnote{Available at https://github.com/basaldella/psytarpreprocessor} and contains train, development and test parts. These parts contain 3535, 431, 1077 entities and 3851, 551, 1192 sentences respectively. We used XLM-RoBERTa-large model  that had been preliminary trained using text data from CommonCrowl project. Fine-tuning of this model provided only for ADR tag excluding WD, SSI, SD tags. The result on the test part was 71.1\% according to the F1 metric achieved with script from the CoNLL evaluation.

\subsection*{Features based on MESHRUS concepts}\label{subsec:meshrus_features}
 MeSH Russian (MESHRUS)~\cite{NLM} is a Russian version of the Medical Subject Headings (MESH) database~\footnote{Home page of the MeSH database site: \url{https://www.nlm.nih.gov/mesh/meshhome.html}}. MESH is a dictionary designed for indexing biomedical information that contains concepts from scientific journal articles and books and is intended for their indexing and searching. The MESH database is filled from articles in English; however, there exist translations of the database to different languages. We used the Russian version, MESHRUS. It is a less complete analogue of the English version, for example, it doesn't contain concept definitions.
MESHRUS contains a set of tuples $(k;v)$ matching Russian concepts $k$ with their relevant CUI codes $v$ from the UMLS thesaurus. A concept $k$ can consist of a word or a sequence of words. 

The following preprocessing algorithm is used
: words are lemmatized, put into a single register and filtered by length, frequency and parts of speech. 
To automatically find and map concepts from MESHRUS to words from corpus we perform two approaches.  

\begin{figure}
    \centering
    \includegraphics[scale=0.27]{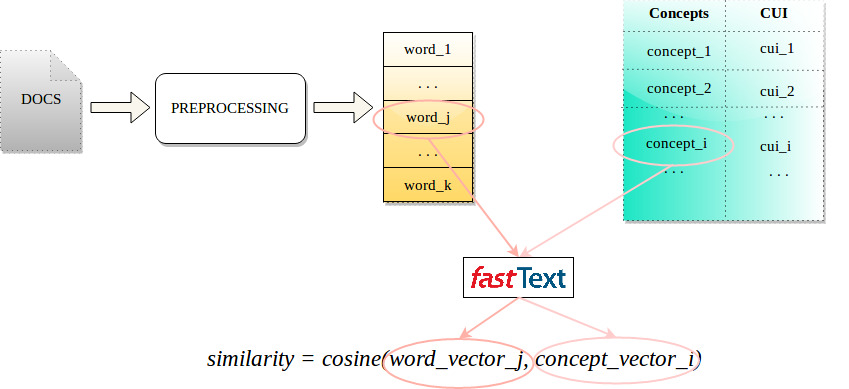}
    \caption{The matching scheme between words of corpus and concepts of UMLS.}
    \label{fig:umls}
\end{figure}

The first approach is to map the filtered words $W=\{w_{i}\}_{i=0}^N$ from the corpus to MESHRUS concepts $\{k_j\}$. As a criterion for comparing words and concepts, we used the cosine similarity between their vector representations obtained using the FastText~\cite{bojanowski2017enriching} model (see Section \ref{subsubsec:features}): a word $w_{i}$ is assigned the CUI code $v_{j}$ (see Fig.~\ref{fig:umls}) whose corresponding concept $k_j$ has the highest similarity measure $\cos\left(\mathrm{FastText}(w_{i}), \mathrm{FastText}(k_{j})\right)$. If this similarity measure is lower than the empirical threshold $T = 0.55$, no CUI code is assigned to $w_i$.



The second approach is based on the mapping of syntactically and lexically related phrases extracted at the sentence level. Prepositions, particles and punctuation are not taken. Syntactic features obtained from dependency trees achieved with UDpipe v2.5. 

For each word $w_{i}\subset W$, its adjacent words $[w_{i-1},w_{i+1}]$ are selected. Together with the word itself they form a lexical set $w_{i_l}$. Then, for the current word $w_{i}$ we find the word $w_{i_\text{parent}}$ that is its parent in the dependency tree (if there is no parent, then the syntactic set contains only $w_{i}$). These $w_{i_l}$ and $w_{i_\text{parent}}$ in turn form a syntactic set $w_{i_s}$. 

Similarly, such lexically and syntactically related sets $c_{j_l}$ and $c_{j_s}$ are formed for each filtered word $c_j$ of the concept from the MESHRUS dictionary: $c_{j_l} = [c_{j-1}, c_{j}, c_{j+1}]$, and $c_{j_s} = [c_{j}, c_{j_\text{parent}}]$.


Further, for each word $w_{i}\subset W$ and word $c_{j}\subset \text{concept}_{k}\subset$ MESHRUS, by analogy with the literature~\cite{shelmanov2015information}, the following metrics are calculated:
\begin{enumerate}
	\item $\mathrm{lexical\_involvement}(w_{i}, c_{j}) = F_1 \left(\frac{|w_{i_l}\cap c_{j_l}|}{|w_{i_l}|}, \frac{|w_{i_l}\cap c_{j_l}|}{|c_{j_l}|} \right)$
	\item $\mathrm{cohesiveness}(w_{i}, c_{j}) = F_1 \left(\frac{|w_{i_s}\cap c_{j_s}|}{|w_{i_s}|}, \frac{|w_{i_s}\cap c_{j_s}|}{|c_{j_s}|} \right)$
    \item $\mathrm{centrality}$ which is 1 if the word $w_{i_\mathrm{parent}}$ of the syntax set $w_{i_s}$ is represented in the syntax set $c_{j_s}$ of words from the dictionary; 0 otherwise.
\end{enumerate}
Here $F_{1}(x, y)$ is the harmonic mean of $x$ and $y$, $|N|$ denotes the length of set $N$, and $M \cap N$ is the intersection of the two sets.
The final metric of similarity between the word $w_i$ and the dictionary concept $c_j$ is calculated as mean of all three metric values.

For each word, its corresponding concept is selected by the highest similarity value provided that the similarity is greater than the specified threshold 0.6. 







\end{document}

%% file: tables/table5.tex
\begin{tabular}{|p{0.2\textwidth}|p{0.755\textwidth}|}
\hline
\textbf{Overall impression}    & ``Помог чересчур!'' (Helped too much!)                                                                                                                                                                                                                                                                                                                                                                                                                                               \\ \hline
\textbf{Advantages}          & ``Цена'' (Price)                                                                                                                                                                                                                                                                                                                                                                                                                                                                   \\ \hline
\textbf{Disadvantages}           & ``отрицательно действует на работоспособность'' (It has a negative effect on productivity)                                                                                                                                                                                                                                                                                                                                                                                         \\ \hline
\textbf{Would you recommend it to friends?} & ``Нет'' (No)                                                                                                                                                                                                                                                                                                                                                                                                                                                                       \\ \hline
\textbf{Comments}             & ``Начала пить недавно. Прочитала отзывы вроде все хорошо отзывались. Стала спокойной даже чересчур, на работе стала тупить, коллеги сказали что я какая то заторможенная, все время клонит в сон. Буду бросать пить эти таблетки.'' (I started taking recently. I read the reviews, and they all seemed positive. I became calm, even too calm, I started to blunt at work, сolleagues said that I somewhat slowed down, feel sleepy all the time. I will stop taking these pills.) \\ \hline
\end{tabular}

%% file: tables/adr_saturation.tex
\renewcommand{\arraystretch}{1.5}
\begin{tabular}{|l|c|c|c|c|c|c|c|c|} 
\hline
\diagbox{\textbf{Parameter}}{\textbf{Corpora}}                                                         &  \textbf{CADEC} & \textbf{TA} & \textbf{TT} & \textbf{TP} & \textbf{N2C2} & \textbf{PSYTAR} & \textbf{RuDRec}  \\ 
\hline
\textbf{total ADR}                                                       & 6318           & 1122        & 899         & 475         & 1579          & 3543            & 720     \\ 
\hline
\textbf{multiword (\%)}                                                 & 72.4           & 0.47        & 40          & 46.7        & 42            & 78              & 54      \\ 
\hline
\textbf{singleword (\%)}                                              & 27.6           & 0.53        & 60          & 53.3        & 58            & 22              & 46      \\ 
\hline
\textbf{discontinuous, non-overlapping (\%)}                              & 1.3            & 0           & 0           & 0           & 0             & 0             & 0       \\ 
\hline
\textbf{continuous, non-overlapping (\%)}                                    & 84             & 100         & 98          & 96.8        & 95            & 100            & 100     \\ 
\hline
\textbf{discontinuous, overlapping (\%)}                                   & 9.3            & 0           & 0           & 0           & 0             & 0             & 0       \\ 
\hline
\textbf{continuous, overlapping (\%)}                                        & 5.3            & 0           & 2           & 3.2         & 5             & 0             & 0       \\ 
\hline
\textbf{$saturation = \frac{Total\ ADR}{number\ of\ words\ in\ corpus}$ $(*10^{3})$}
         &53.38
       &NA & NA 
       &16.5 &1.35       &39.17 &10.61
       \\
\hline
$\frac{Total\ ADR} {Total\ entities\ number}$                             & 0.69           & 0.72        & 0.67        & 0.47        & 0.02          & 0.70            & 0.41    \\ 
\hline
\textbf{$\frac{Total\ ADR}{Number\ of\ Indication,\ Reason,\ etc.}$}          & 22.97          & 7.1         & 1.91          & 0.49           & 0.25          & 0.70          & 0.01       \\ 
\hline
\textbf{Estimation}                                                        & 70.6 \cite{li2020lexicon}         & 61.1 \cite{wang2016mining}      & 64.8  \cite{gupta2018multi}     & 73.6~\cite{li2020exploiting}       & 55.8 \cite{rumshisky2020proceedings}         & \begin{tabular}[x]{@{}c@{}} 71.1 \\ (see Appendix \ref{subsec:psytar_adr}) \end{tabular}          & 60.4~\cite{10.1093/bioinformatics/btaa675}   \\ 
\hline
\textbf{Metric of Estimation}                                          & f1-e           & f1-am       & f1-am       & f1-cs           & f1-r          & f1-e           & f1-e    \\
\hline
\end{tabular}

%% file: tables/types_saturation.tex
\renewcommand{\arraystretch}{1.5}
\begin{tabular}{|p{2.0cm}|p{1.3cm}|p{1.4cm}|p{1.4cm}|p{2.0cm}|p{2.0cm}|p{2.0cm}|p{1.6cm}|c|} 
\hline
\textbf{Entity type}                                                         &  \textbf{Total mentions count} & \textbf{multiword (\%)} & \textbf{singleword (\%)} & \textbf{discontinuous, non-overlapping (\%)} & \textbf{continuous, non-overlapping (\%)} & \textbf{discontinuous, overlapping (\%)} & \textbf{continuous, overlapping (\%)} \\ 
\hline
\textbf{ADR}                                                       & 1784           & 63.85        & 36.15         & 2.97         & 80.66          & 0.62            & 15.75\\ \hline
\textbf{Drugname} & 8236 & 17.13 & 82.87 & 0 & 38.37 & 0.01 & 61.62 \\ \hline
\textbf{DrugBrand} & 4653 & 11.95 & 88.05 & 0 & 0 & 0.02 & 99.98 \\ \hline
\textbf{Drugform} & 5994 & 1.90 & 98.10 & 0 & 83.53 & 0.02 & 16.45 \\ \hline
\textbf{Drugclass} & 3120 & 4.42 & 95.58 & 0 & 94.33 & 0 & 5.67 \\ \hline
\textbf{Dosage} & 965 & 92.75 & 7.25 & 0.10 & 54.92 & 0.21 & 44.77 \\ \hline
\textbf{MedMaker} & 1715 & 32.19 & 67.81 & 0 & 99.71 & 0 & 0.29 \\ \hline
\textbf{Route} & 3617 & 34.95 & 65.05 & 0.53 & 88.80 & 0.06 & 10.62 \\ \hline
\textbf{SourceInfodrug} & 2566 & 48.99 & 51.01 & 6.16 & 91.00 & 0 & 2.84 \\ \hline
\textbf{Duration} & 1514 & 86.53 & 13.47 & 0.20 & 95.44 & 0 & 4.36 \\ \hline
\textbf{Frequency} & 614 & 98.96 & 1.14 & 0.33 & 88.93 & 0 & 10.75 \\ \hline
\textbf{Diseasename} & 4006 & 11.48 & 88.52 & 0.35 & 85.97 & 0.02 & 13.65 \\ \hline
\textbf{Indication} & 4606 & 43.88 & 56.12 & 1.13 & 77.49 & 0.30 & 21.08 \\ \hline
\textbf{BNE-Pos} & 5613 & 66.06 & 33.94 & 1.02 & 82.91 & 0.68 & 15.39 \\ \hline
\textbf{NegatedADE} & 2798 & 92.67 & 7.33 & 1.36 & 87.38 & 0.18 & 11.08 \\ \hline
\textbf{Worse} & 224 & 97.32 & 2.68 & 0.89 & 61.16 & 1.34 & 36.61 \\ \hline
\textbf{ADE-Neg} & 85 & 89.41 & 10.59 & 3.53 & 54.12 & 3.53 & 38.82 \\ \hline
\textbf{Note} & 4517 & 90.21 & 9.79 & 0.13 & 77.77 & 0.15 & 21.94 \\ 
\hline

\end{tabular}

%% file: tables/agreement_t.tex
\begin{tabular}{|p{0.1\textwidth} p{0.1\textwidth}|p{0.08\textwidth}|}
\hline
\multicolumn{1}{|c}{Span strictness, $\alpha$} & \multicolumn{1}{c|}{Tag strictness, $\beta$}            & \multicolumn{1}{c|}{Agreement}                                          

\\ \hline
strict & strict & 61\% \\ 
strict & ignored & 63\% \\ 
intersection & strict & 69\% \\ 
intersection & ignored & 71\% \\ \hline
\end{tabular}

%% file: tables/medication.tex
\begin{tabular}{|p{0.15\textwidth}|p{0.8\textwidth}|}
\hline
\textbf{Drugname}       & Marks a mention of a drug. For example, in the sentence <<Препарат Aventis ``Трентал'' для улучшения мозгового кровообращения>> (The Aventis ``Trental'' drug to improve cerebral circulation), the word ``Trental'' (without quotation marks) is marked as a Drugname.
\\ \hline
\textbf{DrugBrand}      & A drug name is also marked as DrugBrand if it is a registered trademark. For example, in the sentence <<Противовирусный и иммунотропный препарат Экофарм ``Протефлазид''>> (The Ecopharm ``Proteflazid'' antiviral and immunotropic drug), the word ``Протефлазид'' (Proteflazid) is marked as DrugBrand.
\\ \hline
\textbf{Drugform}       & Dosage form of the drug (ointment, tablets, drops, etc.). For example, in the sentence <<Эти таблетки не плохие, если начать принимать с первых признаков застуды>> (These pills are not bad if you start taking them since the first signs of a cold), the word ``таблетки'' (pills) is marked as DrugForm.
\\ \hline
\textbf{Drugclass}      & Type of drug (sedative, antiviral agent, sleeping pill, etc.) For example, in the sentence <<Противовирусный и иммунотропный препарат Экофарм "Протефлазид”>> (The Ecopharm ``Proteflazid'' antiviral and immunotropic drug), two mentions marked as Drugclass: ``Противовирусный'' (Antiviral) and ``иммунотропный'' (immunotropic).
\\ \hline
\textbf{MedMaker}       & The drug manufacturer. This attribute has two values: Domestic and Foreign. For example, in the sentence <<Седативный препарат Материа медика ``Тенотен''>> (The Materia Medica ``Tenoten'' sedative) the word combination ``Материа медика'' (Materia Medica) is marked as MedMaker/Domestic.
\\ \hline
\textbf{MedFrom}        & This is an attribute of a Medication entity that takes one of the two values -- Domestic and Foreign, characterizing the manufacturer of the drug. For example, in the sentence <<Седативные таблетки Фармстандарт “Афобазол”>> (The Pharmstandard ``Afobazol'' sedative pills) the drug name ``Афобазол'' (Afobazol) has its MedFrom attribute equal to Domestic.
\\ \hline
\textbf{Frequency}      & The drug usage frequency. For example, in the sentence <<Неудобство было в том, что его приходилось наносить 2 раза в день>> (Its inconvenience was that it had to be applied two times a day), the phrase ``2 раза в день'' (two times a day) is marked as Frequency.
\\ \hline
\textbf{Dosage}         & The drug dosage (including units of measurement, if specified). For example, in the sentence <<Ректальные суппозитории ``Виферон'' 15000 МЕ -- эффекта ноль>> (Rectal suppositories “Viferon” 150000 IU have zero effect), the mention ``15000 МЕ'' (150000 IU) is marked as Dosage.
\\ \hline
\textbf{Duration}       & This entity specifies the duration of use. For example, in the sentence <<Время использования: 6 лет>> (Time of use: 6 years), ``6 лет'' (6 years) is marked as Duration.
\\ \hline
\textbf{Route}          & Application method (how to use the drug). For example, in the sentence <<удобно то, что можно готовить раствор небольшими порциями>> (it is convenient that one can prepare the solution in small portions), the mention ``можно готовить раствор небольшими порциями'' (can prepare a solution in small portions) is marked as a Route.
\\ \hline
\textbf{SourceInfodrug} & The source of information about the drug. For example, in the sentence <<Этот спрей мне посоветовали в аптеке в его состав входят такие составляющие вещества как мята>> (This spray was recommended to me at a pharmacy, it includes such ingredient as mint), the word combination ``посоветовали в аптеке'' (recommended to me at a pharmacy) is marked as SourceInfoDrug.
\\ \hline
\end{tabular}

%% file: tables/disease.tex
\begin{tabular}{|p{0.15\textwidth}|p{0.8\textwidth}|}
\hline
Diseasename & The name of a disease. If a report author mentions the name of the disease for which they take a medicine, it is annotated as a mention of the attribute Diseasename. For example, in the sentence <<у меня вчера была диарея>> (I had diarrhea yesterday) the word ``диарея'' (diarrhea) will be marked as Diseasename. If there are two or more mentions of diseases in one sentence, they are annotated separately. In the sentence <<Обычно весной у меня сезон аллергии на пыльцу и депрессия>> (In spring I usually have season allergy to pollen, and depression), both ``аллергия'' (allergy) and ``депрессия'' (depression) are independently marked as Diseasename.
\\ \hline
Indication  & Indications for use (symptoms). In the sentence <<У меня постоянный стресс на работе>> (I have a permanent stress at work), the word ``стресс'' (stress) is annotated as Indication. Also, in the sentence <<Я принимаю витамин С для профилактики гриппа и простуды>> (I take vitamin C to prevent flu and cold), the entity ``для профилактики'' (to prevent) is annotated as Indication too. For another example, in the sentence <<У меня температура 39.5>> (I have a temperature of 39,5) the words ``температура 39.5'' (temperature of 39.5) are marked as Indication.
\\ \hline
BNE-Pos     & This entity specifies positive dynamics after or during taking the drug. In the sentence <<препарат Тонзилгон Н действительно помогает при ангине>> (the Tonsilgon N drug really helps a sore throat), the word ``помогает'' (helps) is the one marked as BNE-Pos.
\\ \hline
ADE-Neg     & Negative dynamics after the start or some period of using the drug. For example, in the sentence <<Я очень нервничаю, купила пачку ``персен'', в капсулах, он не помог, а по моему наоборот всё усугубил, начала сильнее плакать и расстраиваться>> (I am very nervous, I bought a pack of ``persen'', in capsules, it did not help, but in my opinion, on the contrary, everything aggravated, I started crying and getting upset more), the words ``по моему наоборот всё усугубил, начала сильнее плакать и расстраиваться'' (in my opinion, on the contrary, everything aggravated, I started crying and getting upset more) are marked as ADE-Neg.                                                                                      \\ \hline
NegatedADE  & This entity specifies that the drug does not work after taking the course. For example, in the sentence <<...боль в горле притупляют, но не лечат, временный эффект, хотя цена великовата для 18-ти таблеток>> (...dulls the sore throat, but does not cure, a temporary effect, although the price is too big for 18 pills) the words ``не лечат, временный эффект'' (does not cure, the effect is temporary) are marked as NegatedADE.                                                                                                                                                                                                                                                                                                     \\ \hline
Worse       & Deterioration after taking a course of the drug. For example, in the sentence <<Распыляла его в нос течении четырех дней, результата на меня не какого не оказал, слизистая еще больше раздражалось>> (I sprayed my nose for four days, it didn't have any results on me, the mucosa got even more irritated), the words ``слизистая еще больше раздражалось'' (the mucosa got even more irritated) are marked as Worse.
\\ \hline
\end{tabular}

%% file: tables/general_inf_about_corpus_2.tex
\begin{tabular}{|l|l|p{0.13\textwidth}|p{0.12\textwidth}|p{0.075\textwidth}|p{0.07\textwidth}} 
\hline
\multirow{2}{*}{Entity type} & \multicolumn{4}{c|}{Mentions}                               \\ 
\cline{2-5}
                                                                                   & Annotated   & Classification \& normalization & Num. words in the mentions & Reviews coverage \\ 
\hline
ADR & 1784 & 316 (MedDRA) & 4 211 & 628 \\ 
\hline
Medication                                                                         & 32 994 &    & 47 306 & 2 799  \\ 
\hline
Drugname                                                                           & 8 236  & 550(SRD), 226(ATC)    & 9 914                         & 2 793  \\ 
\hline
DrugBrand                                                                          & 4 653  &     & 5 296                        & 1 804 \\ 
\hline
Drugform & 5 994  &  & 6 131 & 2 193  \\ 
\hline
Drugclass & 3 120  & 70 (ATC) & 3 277 & 1 687 \\ 
\hline
MedMaker                                                                           & 1 715    &     & 2 423 & 1 448 \\ 
\hline
Frequency                                                                          & 614    &     & 2 478 & 516 \\ 
\hline
Dosage                                                                             & 965    &      & 2 389 & 708 \\ 
\hline
Duration                                                                           & 1 514    &      & 3 137 & 1 194  \\ 
\hline
Route                                                                              & 3 617  &     & 7 869                           & 1 737  \\ 
\hline
SourceInfodrug                                                                     & 2 566  &       & 4 392 & 1 579   \\ 
\hline
Disease                                                                            & 17 332  &    & 37 863 & 2 712  \\ 
\hline
Diseasename                                                                        & 4 006  & 247 (ICD-10)    & 4 713 & 1 621  \\ 
\hline
Indication & 4 606  & 343 (MedDRA) & 7 858 & 1 784  \\ 
\hline
BNE-Pos                                                                            & 5 613  &    & 14 883 & 1 764  \\ 
\hline
ADE-Neg                                                                            & 85    &      & 347 & 54 \\ 
\hline
NegatedADE                                                                         & 2 798  &     & 9 028 & 1 104  \\ 
\hline
Worse                                                                              & 224     &      & 1 034 & 134  \\ 
\hline
Note                                                                               & 4 517  &    & 21 200 & 1 876  \\
\hline
\end{tabular}

%% file: tables/coref_number_comparison.tex
\begin{tabular}{|p{0.1\textwidth}| p{0.08\textwidth}|p{0.08\textwidth}|p{0.08\textwidth}|}
\hline
Corpus &  Texts count &  Mentions count &  Chains count 
\\ \hline
AnCor-2019 & 522 & 25159 &5678 \\ \hline
Our corpus & 300 & 6276 & 1560 \\ \hline
\end{tabular}

%% file: tables/coref_medtypes.tex
\begin{tabular}{|l|l|p{0.18\textwidth}|}
\hline
Entity type &  Attribute type & Number of mentions involved in coreference chains 

\\ \hline
\multirow{8}{*}{Medication} & Drugname & 529  \\ \cline{2-3}
 & Drugform & 286 \\ \cline{2-3}
& Drugclass & 204 \\ \cline{2-3}
& MedMaker & 170 \\ \cline{2-3}
& Route & 98 \\ \cline{2-3}
& SourceInfodrug & 75 \\ \cline{2-3}
 & Dosage & 50 \\ \cline{2-3}
& Frequency & 1 \\ \hline
\multirow{6}{*}{Disease}  & Diseasename & 163 \\  \cline{2-3}
 & Indication & 125 \\ \cline{2-3}
 & BNE-Pos & 107 \\ \cline{2-3}
 & NegatedADE & 36 \\ \cline{2-3}
 & Worse & 5 \\ \cline{2-3}
 & ADE-Neg & 2 \\ \hline
\multicolumn{2}{| c |}{ADR} & 34 \\ \hline
\end{tabular}

%% file: tables/embedding_exp.tex

\centering
\begin{tabular}{|c|c|l|l|l|} 
\hline
\begin{tabular}[c]{@{}c@{}}\textbf{Word vector }\\\textbf{ representation }\end{tabular} & \begin{tabular}[c]{@{}c@{}}\textbf{Vector }\\\textbf{ dimension }\end{tabular} & \multicolumn{1}{c|}{\textbf{ADR }} & \multicolumn{1}{c|}{\textbf{Medication }} & \multicolumn{1}{c|}{\textbf{Disease }}  \\ 
\hline
FastText                                                                                 & 300                                                                            & 22.4 $\pm$ 1.6                     & 70.4 $\pm$ 1.1                            & 44.1 $\pm$ 1.7                          \\ 
\hline
ELMo                                                                                     & 1024                                                                           & 24.3 $\pm$ 1.7                     & 73.4 $\pm$ 1.5                            & 46.4 $\pm$ 0.6                          \\ 
\hline
BERT                                                                                     & 768                                                                            & 22.1 $\pm$ 2.4                     & 71.4 $\pm$ 3.3                            & 45.5 $\pm$ 3.2                          \\ 
\hline
\multicolumn{1}{|l|}{ELMO\textbar{}\textbar{}BERT}                                       & \multicolumn{1}{l|}{1024\textbar{}\textbar{}768}                               & 18.7 $\pm$ 9.8                     & 74.1 $\pm$ 1.1                            & 47.9 $\pm$ 1.6                          \\
\hline
\end{tabular}

%% file: tables/feature_top_exp.tex
\begin{tabular}{|c|c|c|c|}
\hline
\textbf{\textbf{Topology and features}}                                                                & \textbf{\textbf{ADR}}               & \textbf{\textbf{Medication}} & \textbf{\textbf{Disease}} \\ \hline
\multicolumn{4}{|c|}{\textbf{Model A - Influence of features}}                                                                                                                                                                  \\ \hline
ELMo + PoS                                                                                             & 26.2 $\pm$ 3.0                      & 72.9 $\pm$ 0.6               & 46.6 $\pm$ 0.9            \\ \hline
ELMo + ton                                                                                             & 26.6 $\pm$ 3.9                      & 73.5 $\pm$ 0.5               & 47.3 $\pm$ 1.0            \\ \hline
ELMo + Vidal                                                                                           & 26.8 $\pm$ 1.0                      & 73.2 $\pm$ 1.1               & 45.8 $\pm$ 1.2            \\ \hline
ELMo + MESHRUS                                                                                         & 27.4 $\pm$ 2.2                      & 73.3 $\pm$ 1.5               & 46.5 $\pm$ 1.2            \\ \hline
ELMo + MESHRUS-2                                                                                       & 27.4 $\pm$ 0.9                      & 73.1 $\pm$ 0.4               & 46.7 $\pm$ 1.4            \\ \hline
\multicolumn{4}{|c|}{\textbf{Model A - Topology modifications}}                                                                                                                                         \\ \hline
ELMo, 3-layer LSTM                                                                                     & 28.2 $\pm$ 5.1                      & 74.7 $\pm$ 0.7               & 51.5 $\pm$ 1.8            \\ \hline
ELMo, CRF                                                                                              & 28.8 $\pm$ 2.7                      & 73.2 $\pm$ 1.1               & 46.9 $\pm$ 0.4            \\ \hline
\multicolumn{4}{|c|}{\textbf{Model A - Best combination}}                                                                                                                                               \\ \hline
\begin{tabular}[c]{@{}c@{}}ELMo, 3-layer LSTM, CRF \\ ton, PoS, MESHRUS, MESHRUS-2, Vidal\end{tabular} & 32.4 $\pm$ 4.7                      & 74.6 $\pm$ 1.1               & 52.3 $\pm$ 1.4            \\ \hline
\multicolumn{4}{|c|}{\textbf{Model B - XLM-RoBERTa part only}}                                                                                                                                          \\ \hline
\multicolumn{1}{|l|}{XLM-RoBERTa-large}                                                                & \multicolumn{1}{l|}{40.1 $\pm$ 2.9} & \multicolumn{1}{l|}{79.6 $\pm$ 1.3}      & \multicolumn{1}{l|}{56.9 $\pm$ 0.8}   \\ \hline
\end{tabular}

%% file: tables/rdrs_subsets.tex
\setlength\tabcolsep{2.2pt}
\begin{tabular}{|p{3.25cm}|c|c|c|c|c|c|} 
\hline
\diagbox{\textbf{Parameters}}{\textbf{Corpora}} & \textbf{RDRS 2800}   & \textbf{RDRS 1600}  & \textbf{RDRS 1250}  & \textbf{RDRS 610}   & \textbf{RDRS 1136}  & \textbf{RDRS 500}   \\ 
\hline
\textbf{Number of reviews }                         & 2800            & 1659           & 1250           & 610            & 1136           & 500 \\ 
\hline
\textbf{Number of reviews contained ADR }           & 625             & 339            & 610            & 610            & 610            & 177 \\ 
\hline
\textbf{Portion of reviews contained ADR }          & 0.22            & 0.2            & 0.49           & 1              & 0.54           & 0.35 \\ 
\hline
\textbf{Number of ADR entitites }                   & 1778            & 843            & 1752           & 1750           & 1750           & 709  \\ 
\hline
\textbf{Average number of ADR per review }          & 0.64            & 0.51           & 1.4            & 2.87           & 1.54           & 1.42   \\ 
\hline
\textbf{Number of reviews contained Indication }    & 1783            & 955            & 670            & 59             & 154            & 297  \\ 
\hline
\textbf{Total entitites number }                    & 52186           & 27987          & 21807          & 3782           & 6126           & 9495    \\ 
\hline
\textbf{Number of Indication entitites }            & 4627            & 2310           & 1518           & 90             & 237            & 720    \\ 
\hline
\textbf{Portion of ADR to Indication entities }     & 0.38            & 0.36           & 1.15           & 19.44          & 7.38           & 0.98      \\ 
\hline
\textbf{F1-exact }                                  & $52.8\pm 3.8$ & $40.1\pm2.9$ & $61.1\pm1.5$ & $71.3\pm3.4$ & $68.6\pm3.3$ & $61.6\pm2.9$  \\ 
\hline
\textbf{Saturation $(*10^{3})$}                                 & 4.25        & 3.41       & 9.77       & 72.57       & 42.99       & 9.08       \\\hline
\end{tabular}

%% file: tables/all_tags_2800_model_B.tex
\begin{tabular}{|l|c|c|} 
\hline
\diagbox{\textbf{Entity type}}{\textbf{Corpora}}     & \textbf{RDRS 1250} & \textbf{RDRS 2800}  \\ 
\hline
BNE-Pos        & 51.2                 & 50.3             \\ 
\hline
Diseasename    & 87.6                 & 88.3             \\ 
\hline
Indication     & 58.8                 & 62.2             \\ 
\hline
MedFromDomestic       & 61.7                 & 76.2             \\ 
\hline
MedFromForeign        & 63.5                 & 74.4             \\ 
\hline
MedMakerDomestic      & 65.1                 & 87.1             \\ 
\hline
MedMakerForeign       & 74.4                 & 85.0             \\ 
\hline
Dosage         & 59.6                 & 63.2             \\ 
\hline
DrugBrand      & 81.5                 & 83.8             \\ 
\hline
Drugclass      & 89.7                 & 90.4             \\ 
\hline
Drugform       & 91.5                 & 92.4             \\ 
\hline
Drugname       & 94.2                 & 95.0            \\ 
\hline
Duration       & 75.5                 & 74.7             \\ 
\hline
Frequency      & 63.4                 & 65.0             \\ 
\hline
MedMaker       & 92.5                 & 93.8             \\ 
\hline
Route          & 58.4                 & 61.2             \\ 
\hline
SourceInfodrug & 66.0                 & 67.3             \\
\hline
Negative* & 52.2 & 52.0 \\
\hline
\end{tabular}

%% file: tables/integral_scores.tex
\setlength\tabcolsep{2.2pt}
\begin{tabular}{|l|c|c|c|}
\hline
                     & \multicolumn{3}{c|}{\textbf{F1 exact}}                                                            \\
\hline
\textbf{RDRS subset} & \textbf{ADR}                                            & \textbf{Disease} & \textbf{Medication}  \\
\hline
2800                 & $52.8\pm3.4$ & $63.5\pm0.5$    & $84.1\pm0.8$        \\
\hline
1250                 & $61.1\pm1.5$  & $62.9\pm1.5$   & $84.2\pm0.6$                 \\
\hline
1600                 & $40.1\pm2.7$ & $56.9\pm0.9$    & $79.6\pm1.3$        \\
\hline
\end{tabular}

%% file: tables/coref_eval.tex
\begin{tabular}{|p{0.14\textwidth}|p{0.14\textwidth}|p{0.06\textwidth}|p{0.06\textwidth}|p{0.07\textwidth}|p{0.08\textwidth}|}
\hline
Training corpus &  Testing corpus &  avg F1 &  $B^3$ F1 & MUC F1 & CEAFe F1
\\ \hline
AnCor-2019 & RDR & 58.7 & 56.4 & 61.3 & 58.3 \\ \hline
AnCor-2019 & AnCor-2019 & 58.9 & 55.6 & 65.1 & 55.9 \\ \hline
Our corpus  & Our corpus  & 71.0 & 69.6 & 74.2 & 69.3 \\ \hline
Our corpus  & AnCor-2019 & 28.7 & 26.5 & 33.3 & 26.4 \\ \hline
AnCor-2019 + Our corpus  & Our corpus  & 49.4 & 47.6 & 52.2 & 48.4 \\ \hline
AnCor-2019 + Our corpus & AnCor-2019 & 31.8 & 31.4 & 40.7 & 23.3 \\ \hline
\end{tabular}